\documentclass[review,5p,twocolumn]{elsarticle}
\usepackage{times}
\usepackage{booktabs}
\usepackage{multirow}
\usepackage{lineno,hyperref}
\usepackage{epsfig}
\usepackage{graphicx}
\usepackage{amsmath}
\usepackage{amssymb}
\usepackage{algorithm}
\usepackage{algorithmicx}
\usepackage{algpseudocode}
\usepackage{balance}
\usepackage{CJK}
\usepackage{acronym}
\usepackage{color}
\usepackage{colortbl}
\usepackage{array}
\usepackage{graphicx}

\journal{Journal of Pattern Recognition}
\biboptions{numbers,sort&compress}
\modulolinenumbers[5]

\begin{document}
\begin{frontmatter}

\title{Deep Self-Paced Learning for Person Re-Identification}
\author[1]{Sanping Zhou}
\ead{sanpingzhou@stu.xjtu.edu.cn}
\author[1]{Jinjun Wang\corref{cor1}}
\cortext[cor1]{Corresponding author:
  Tel.: +86-029-83395146;
  Fax: +86-029-83395175;}
\author[2]{Deyu Meng}
\author[1]{Xiaomeng Xin}
\author[3]{Yubing Li}
\author[1]{Yihong Gong}
\author[1]{Nanning Zheng}
\address[1]{The institute of artificial intelligence and  robotic, Xi'an Jiaotong University, Xian Ning West Road No.28, Shaanxi, 710049, P.R. China}
\address[2]{School of Mathematics and Statistics, Xi'an Jiaotong University, Xian Ning West Road No.28, Shaanxi, 710049, P.R. China}
\address[3]{School of the Electronic and Information Engineering, Xi'an Jiaotong University, Xian Ning West Road No.28, Shaanxi, 710049, P.R. China}

\begin{abstract}
Person re-identification~(Re-ID) usually suffers from noisy samples with background clutter and mutual occlusion, which makes it extremely difficult to distinguish different individuals across the disjoint camera views. In this paper, we propose a novel deep self-paced learning~(DSPL) algorithm to alleviate this problem, in which we apply a self-paced constraint and symmetric regularization to help the relative distance metric training the deep neural network, so as to learn the stable and discriminative features for person Re-ID. Firstly, we propose a soft polynomial regularizer term which can derive the adaptive weights to samples based on both the training loss and model age. As a result, the high-confidence fidelity samples will be emphasized and the low-confidence noisy samples will be suppressed at early stage of the whole training process. Such a learning regime is naturally implemented under a self-paced learning~(SPL) framework, in which samples weights are adaptively updated based on both model age and sample loss using an alternative optimization method. Secondly, we introduce a symmetric regularizer term to revise the asymmetric gradient back-propagation derived by the relative distance metric, so as to simultaneously minimize the intra-class distance and maximize the inter-class distance in each triplet unit. Finally, we build a part-based deep neural network, in which the features of different body parts are first discriminately learned in the lower convolutional layers and then fused in the higher fully connected layers. Experiments on several benchmark datasets have demonstrated the superior performance of our method as compared with the state-of-the-art approaches.
\end{abstract}


\end{frontmatter}

\section{Introduction}
\label{sec_intr}
Person re-identification~(Re-ID) has become an active research topic in the field of computer vision, because of its wide application in the video surveillance community. Given one single shot or multiple shots of a target, person Re-ID concerns the problem of matching the same person among a set of gallery candidates captured from the disjoint camera views~\cite{Zhou_Wang_Wang:2017,Zhong_Zheng_Cao:2017,Su_Zhang_Yang:2017,Liu_Wang_Wang:2017}. It is a very challenging task due to noisy samples with mutual occlusion and background clutter that makes the large appearance variations across different camera views~\cite{Li_Chang_Wang:2015,Ren_Lu_Feng:2017}. Therefore, the key to improve the identification performance is to learn the stable and discriminative features for representation.

The fundamental person Re-ID problem is to compare an image of each interested target seen in a probe camera view to a large number of candidates captured from a gallery camera view which has no overlap with the probe one~\cite{Ma_Zhu_Gong:2017}. If a true match to the probe exists in the gallery, it should have a higher similarity score as compared with the incorrect matches. Previous efforts for solving this problem primarily focus on the following two aspects: 1) developing robust feature descriptors to handle the variations in person's appearance, and 2) designing discriminative distance metrics to measure the similarity of person's images. For the first category, different cues are employed for the stable and discriminative features. Representative descriptors include the Local Binary Pattern~(LBP)~\cite{Xiong_Gou_Camps:2014}, Ensemble of Local Feature~(ELF)~\cite{Gray_Tao:2008} and Local Maximal Occurrence~(LOMO)~\cite{Zhao_Ouyang_Wang:2014}. For the second category, labeled images are used to train a distance metric, in which the intra-class distance is minimized while the inter-class distance is maximized. Typical metric learning methods include the Locally Adaptive Decision Function~(LADF)~\cite{Li_Chang_Liang:2013}, Large Margin Nearest Neighbor~(LMNN)~\cite{Weinberger_Blitzer_Saul:2006} and Information Theoretic Metric Learning~(ITML)~\cite{Davis_Kulis_Jain:2007}. Since both line of works regard the feature extraction and metric learning processes as two disjoint steps, their performances are limited.

In the past two years, the deep convolutional neural network ~(CNN) based methods~\cite{Ahmed_Jones_Marks:2015,Ding_Lin_Wang:2015,Zhou_Wang_Hou:2016, Zheng_Zheng_Yang:2016,Zhou_Wang_Shi:2017,Wu_Shen_Van:2017} have been proposed to combine the feature extraction and metric learning into an end-to-end learning framework, in which a neural network is built to extract the stable and discriminative features under the supervision of a suitable distance metric. Benefit from the powerful representation capability of the deep CNN, this line of methods have achieved promising results on the benchmark datasets for person Re-ID. The relative distance metric~\cite{Wang_Song_Leung:2014} has been widely used as loss function in the deep learning based methods for visual recognition. Compared with the well-known softmax loss~\cite{Krizhevsky_Sutskever_Hinton:2012}, it is a better choice for the zero-shot recognition problem, because of the training set doesn't have the same identity with the testing set. The relative distance metric aims to maximize the relative distance between the positive pair and negative pair in each triplet unit, which can generate a large number of triplet inputs even using a small number of training samples.  Therefore, it is very suitable choice for the person Re-ID problem which not only is a zero-shot problem but also can only provide the small-scale dataset for training.

\begin{figure}[!htb]
\footnotesize
\centering
    \begin{tabular}{c}
        \hspace{-0.4cm}
        \includegraphics[height = 5.0cm, width = 9.0cm]{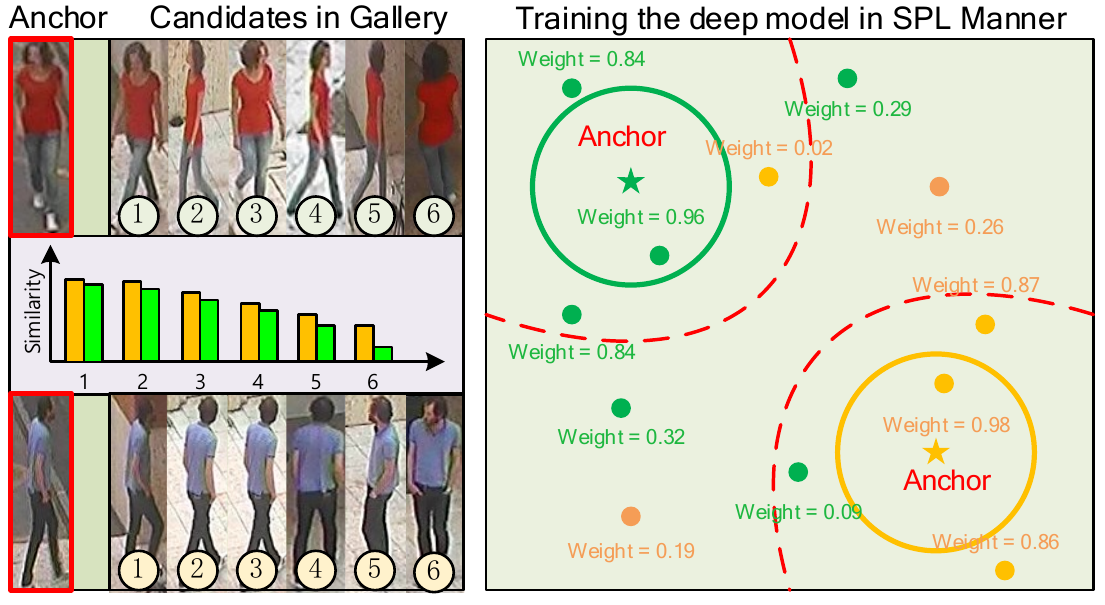}
    \end{tabular}
    \vspace{-0.3cm}
    \caption{Illustration of our SPL motivations in dealing with the noisy training samples or outliers. The left column shows some typical positive candidates to two anchor images, in which the similarity scores of these positive candidates to the anchor vary from large to small with the incensement of indexes. The right column shows the SPL training strategy , in which the derived weighting scheme will adaptively update the sample weights according to the training loss and model age. Therefore, high-confidence fidelity samples will be emphasized and the the low-confidence noisy samples will be suppressed at early stage of the whole learning process.}
    \label{fig_1}
\end{figure}

To further improve the identification performance, our observation shows that the following three issues should also be addressed in the learning process. Firstly, the order and weight of training samples should be considered, as shown in Fig.~\ref{fig_1}, otherwise it might be easy to cause the unstable learning due to the noisy samples or outliers with mutual occlusion and background clutter. Secondly, it is unsuitable to directly apply the distance metric to supervise the training process of deep CNN without any regularization to the gradient back-propagation. Because most of the deep learning tools, such as Caffe~\cite{Jia_Shelhamer_Donahue:2014} and Tensorflow~\cite{Abadi_Agarwal_Barham:2016}, take the gradient back-propagation algorithm to optimize the deep parameters. Thirdly, the neural network should be relatively small and include the part processing module, due to the person Re-ID is a fine-grained problem and the dataset for person Re-ID is usually in small size. As a consequence, it is very  urgent to study the three aspects of problems in the training process.

In this paper, we propose a novel deep self-paced learning ~(DSPL) algorithm to adaptively update the weights to samples and regularize the gradient back-propagation of relative distance metric~\cite{Wang_Song_Leung:2014} in the learning process, so as to further improve the identification performance of deep neural network for person Re-ID. In order to extract the stable and discriminative features, we firstly build a part-based deep neural network, in which the features of different body parts are discriminately learned in the lower convolutional layers and then fused in the higher fully connected layers. Then, we introduce the self-paced learning~(SPL) theory~\cite{Kumar_Packer_Koller:2010} into the training framework, in which samples can be ranked in a self-paced manner by applying a novel soft polynomial regularizer term to adaptively update the weights according to both the model age and sample loss in each iteration. Specially, the high-confidence fidelity samples will be emphasized and the the low-confidence noisy samples will be suppressed at early stage of the whole learning process. Therefore, the neural network can be trained in a stable process by gradually involving the faithful samples from easy to hard. In addition, a symmetric regularizer term is introduced to overcome the drawback of relative distance metric in gradient back-propagation. As a result, the intra-class distance is minimized and the inter-class distance is maximized by regularizing the asymmetric gradient back-propagation in each triplet unit. Extensive experimental results on several benchmark datasets have shown that our method performs much better than the state-of-the-art approaches.

In summary, the main contributions of this paper can be highlighted as follows:
\begin{itemize}
  \item We propose a novel DSPL algorithm to supervise the learning of deep neural network, in which a soft polynomial regularizer term is proposed to gradually involve the faithful samples into training process in a self-paced manner.
  \item We optimize the gradient back-propagation of relative distance metric by introducing a symmetric regularizer term, which can convert the back-propagation from the asymmetric mode to a symmetric one.
  \item We build an effective part-based deep neural network, in which features of different body parts are first discriminately learned in the lower convolutional layers and then fused in the higher fully connected layers.
\end{itemize}

The rest of our paper is organized as follows: Section~\ref{sec_related} reviews some of the related works. In Section~\ref{sec_method}, we describe the proposed method, including the DSPL algorithm and deep neural network. The experimental results and corresponding analysis are presented in Section~\ref{sec_experiment}. Conclusion comes in Section~\ref{sec_conclusion}.

\section{Related work}
\label{sec_related}
In this section, we review two lines of related works, namely the \emph{Person Re-ID} and \emph{Self-Paced Learning}, which are briefly introduced in the following paragraphs.
\subsection{Person Re-ID}
Extensive works have been reported to address the person Re-ID problem, which mainly focus on several aspects of the issue, such as developing robust feature descriptors, designing distinctive distance metrics and learning stable and discriminative deep features for representation. Blew we will give a brief review of some representative ones.

The feature designing based methods mainly focus on developing robust feature descriptors which are invariant to the view angles, lighting conditions, body poses and background clutters. For example, Zhao et al.~\cite{Zhao_Ouyang_Wang:2014} learned a mid-level filter from patch cluster to achieve cross-view invariance. In~\cite{Liao_Hu_Zhu:2015}, Liao et al. constructed a feature descriptor which analyzed the horizontal occurrence of local features and maximized the occurrence to obtain a robust feature representation against viewpoint changes. Ma et al.~\cite{Ma_Su_Jurie:2012} presented the person image via covariance descriptor which was robust to illumination changes and background variations. In~\cite{Farenzena_Bazzani_Perina:2010}, Farenzena et al. augmented maximally stable color regions with histograms for person representation. Zhao et al.~\cite{Zhao_Ouyang_Wang:2017} learned the distinct salience features to distinguish the matched person from others. In~\cite{Cheng_Cristani_Stoppa:2011}, Chen et al. employed a pre-learned pictorial structure model to localize the body parts more accurately. Wu et al.~\cite{Wu_Li_Radke:2015} introduced a viewpoint invariant descriptor, which took the viewpoint of the human into account by using what they called a pose prior learned from the training data. In~\cite{Koestinger_Hirzer_Wohlhart:2012}, Kviatkovsky et al. investigated the intra-distribution structure of color descriptor, which was invariant under certain illumination changes. Li et al.~\cite{Li_Wang:2013} matched person images observed in different camera views with complex cross-view transformations and applied it to the person Re-ID problem. These methods aim to improve the person Re-ID performance by developing a fixed feature descriptor, however the adaptive feature learning is not addressed.

The metric learning based methods aim to find a mapping function from the feature space to another distance space where feature vectors from the same person are more similar than those from different ones. For example, Zheng et al.~\cite{Zheng_Gong_Xiang:2013} proposed a relative distance learning method from the probabilistic prospective. In~\cite{Mignon_Jurie:2012}, Mignon et al. learned a distance metric from the sparse pairwise similarity constraints. Pedagadi et al.~\cite{Pedagadi_Orwell_Velastin:2013} utilized the LADF to map high dimensional features into a more discriminative low dimensional space. In~\cite{Xiong_Gou_Camps:2014}, Xiong et al. further extended the LADF and several other metric learning methods by using kernel tricks and different regularizers. Nguyen et al.~\cite{Nguyen_Bai:2011} measured the similarity of face pairs through the cosine similarity, which is closely related to the inner product similarity. In~\cite{Loy_Liu_Gong:2013}, Loy et al. casted the person Re-ID problem as an image retrieval task by considering the listwise similarity. Chen et al.~\cite{Chen_Yuan_Hua:2017} proposed a kernel based metric learning method to explore the nonlinearity relationship of samples in the feature space. In~\cite{Hirzer_Roth:2012}, Hirzer et al. learned a discriminative distance metric by using the relaxed pairwise constraints. Prosser et al.~\cite{Prosser_Zheng_Gong:2010} developed a ranking model using a support vector machine. These methods learn a specific distance metric mainly based on feature representation extracted by several fixed feature descriptors, which may influence the performance of metric learning.

Different from the above mentioned two lines of methods, the deep learning based methods usually incorporate the feature extraction and metric learning into an end-to-end learning framework, in which a deep neural network is built to extract features from the input images and a distance metric is used to compute the loss and back-propagate the gradients. For example, Ahmed et al.~\cite{Ahmed_Jones_Marks:2015} proposed a novel deep neural network which took the pairwise images as inputs, and outputted a similarity score indicating whether the two input images were the same person or not. In~\cite{Xiao_Li_Ouyang:2016}, Xiao et al. applied a domain guided dropout algorithm to improve the performance of deep CNN in extracting general feature representations. Ding et al.~\cite{Ding_Lin_Wang:2015} introduced a triplet neural network to learn the relative similarity under supervision of the triplet loss. In~\cite{Wang_Zuo_Lin:2016}, Wang et al. proposed a unified triplet loss and siamese deep architecture, which can jointly extract single-image and cross-image feature representations. Zhang et al.~\cite{Zhang_Lin_Zhang:2015} incorporated the deep hash learning into a triplet formulation and efficiently improved the identification speed. In~\cite{Yi_Lei_Liao:2014}, Yi et al. constructed a siamese architecture to learn pairwise similarity and used body part strategy to design the neural network. Li et al.~\cite{Li_Zhao_Xiao:2014} proposed a novel filter pairing neural network to model body part displacements by using the patch matching layers to match the filter responses of local patches. In~\cite{Chen_Zhu_Zheng:2017}, Chen et al. learned a view-specific feature transformation by considering the camera correlation in the deep learning framework. Yan et al.~\cite{Yan_Ni_Song:2016} proposed a progressive fusion framework based on the LSTM, so as to aggregate the frame-wise human region representation and yield a sequence level feature representation for person Re-ID. These methods usually incorporate the feature extraction and metric learning into a joint framework mainly based on the general neural networks, such as AlexNet~\cite{Krizhevsky_Sutskever_Hinton:2012} and VGGNet~\cite{Simonyan_Zisserman:2014}, without applying an effective part strategy in the neural networks, which may be inappropriate for the person Re-ID problem.

\begin{figure*}[!htb]
\footnotesize
\centering
    \begin{tabular}{c}
        \includegraphics[height = 5.3cm, width = 17.5cm]{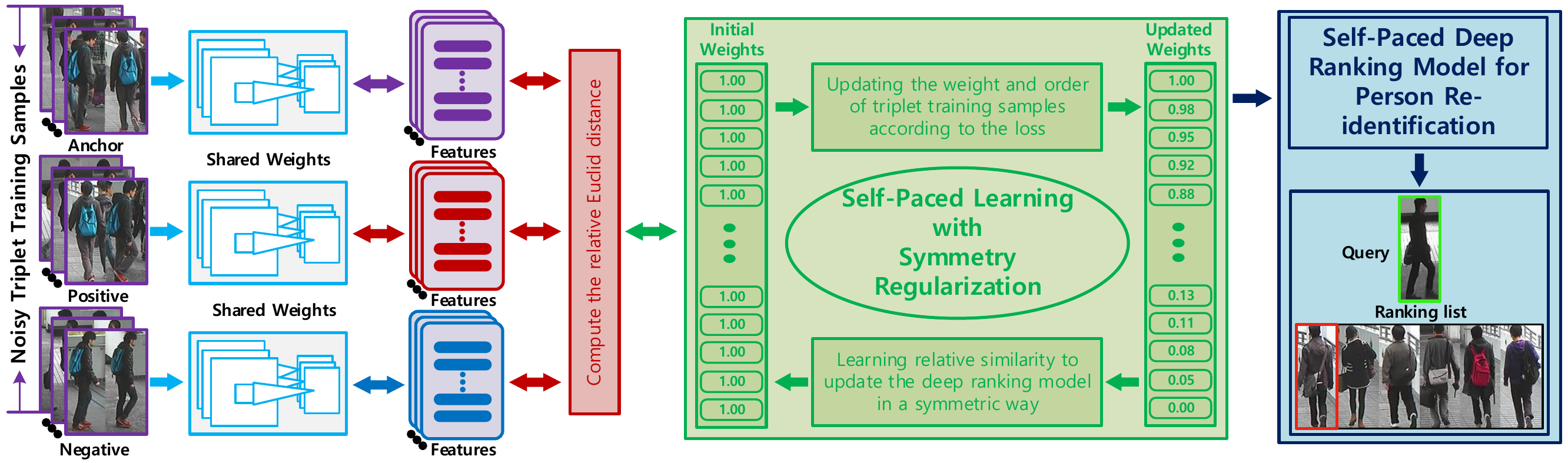}
    \end{tabular}
    \caption{The framework of our deep self-paced person re-identification method. Take the rare images as inputs, our method can effectively alleviate the side effects of noisy training samples or outliers by imposing adaptive weights on them in the relative similarity comparison framework. The SPL constraint gives smaller weights to the noisy low-confidence samples and larger weights to the clean high-confidence samples. As a result, the generalization ability of neural network can be gradually strengthened to deal with the cross-view appearance variations.}
    \label{fig_2}
\end{figure*}

\subsection{Self-Paced Learning}
The SPL theory is inspired by the cognitive process of human beings, where samples are involved into the training process from easy to hard ones~\cite{Bengio_Louradour_Collobert:2009}. As an effective strategy to suppress the side effects of noisy samples or outliers, the SPL based methods have been witnessed the great successes in various machine learning fields~\cite{Kumar_Packer_Koller:2010}. For example, Jiang et al.~\cite{Jiang_Meng_Yu:2014} incorporated the diversity concept into a SPL framework to deal with the event detection and action recognition problem. In~\cite{Zhang_Meng_Li:2017}, Zhang et al. integrated the multiple-instance learning problem into a SPL regime, so as to improve the performance in co-saliency detection. Lee and Grauman~\cite{Lee_Grauman:2011} proposed a self-paced approach to gradually learn from the complex samples in visual category discovery. In~\cite{Supancic_Ramanan:2013}, Supancic and Ramanan applied the SPL theory to choose an appropriate framework to learn good appearance model for long-term tracking. Tang et al.~\cite{Tang_Ramanathan_Li:2012} proposed a self-paced domain adaptation method to adapt a object detector from the image domain to the video domain. In~\cite{Li_Gong_Meng:2016}, Li et al. proposed a multi-objective method to enhance the convergence of the SPL based algorithms. Zhao et al.~\cite{Zhao_Meng_Jiang:2015} proposed a novel matrix factorization learning methodology by introducing a soft self-paced regularizer term to impose adaptive weights to samples. In~\cite{Liang_Li_Cao:2016}, Liang et al. proposed a self-paced cross modal subspace matching method which can gradually chooses the faithful samples to train the model by updating weights in a self-paced manner. Lin et al.~\cite{Lin_Wang_Meng:2017} developed a novel cost-effective framework to deal with the face identification problem, which utilized the high-confidence and low-confidence samples in both the self-paced and active user-query way. These methods mainly apply the SPL theory in the traditional metric learning framework, and we do not see its application in the popular deep learning framework.
\section{The proposed method}
\label{sec_method}

Let $\mathbf{X} = \{\mathbf{X}_i | \left(\mathbf{x}_i^a, \mathbf{x}_i^p, \mathbf{x}_i^n\right)\}_{i=1}^N$ be the triplet training units, where $\{\mathbf{x}_i^a, \mathbf{x}_i^p\}$ denotes the positive pair, $\{\mathbf{x}_i^a, \mathbf{x}_i^n\}$ represents the negative pair and $N$ denotes the number of total triplet units. The goal of our deep architecture is to learn the filter weights and biases that minimizes the ranking error from the output layer. A recursive function for an $K$-layer deep model can be formulated as follows:
\begin{equation}
\label{eq_1}
    \begin{aligned}
        \mathbf{X}_i^{k} &= \Psi(\mathbf{W}^{k}*\mathbf{X}_i^{k-1} + \mathbf{b}^{k}),\\
        i = 1, 2, \cdots&, N; k = 1, 2, \cdots, K; \mathbf{X}_i^{0} = \mathbf{X}_i.
    \end{aligned}
\end{equation}
where $\mathbf{W}^{k}$ denotes the filter weights of the $k^{th}$ layer, $\mathbf{b}^{k}$ refers to the corresponding biases, $*$ denotes the convolution operation, $\Psi(\cdot)$ is an element-wise non-linear activation function such as ReLU, and $\mathbf{X}_i^{k}$ represents the feature maps generated at layer $k$ for sample $\mathbf{X}_i$. For simplicity, we simplify the parameters of the neural network as a whole and define $\mathbf{W} = \{\mathbf{W}^{1}, \cdots, \mathbf{W}^{K}\}$ and $\mathbf{b} = \{\mathbf{b}^{1}, \cdots, \mathbf{b}^{K}\}$.

\subsection{Deep self-paced person Re-ID}

\begin{figure}[!htb]
\footnotesize
\centering
    \begin{tabular}{c}
        \hspace{-0.3cm}
        \includegraphics[height = 4.8cm, width = 9.0cm]{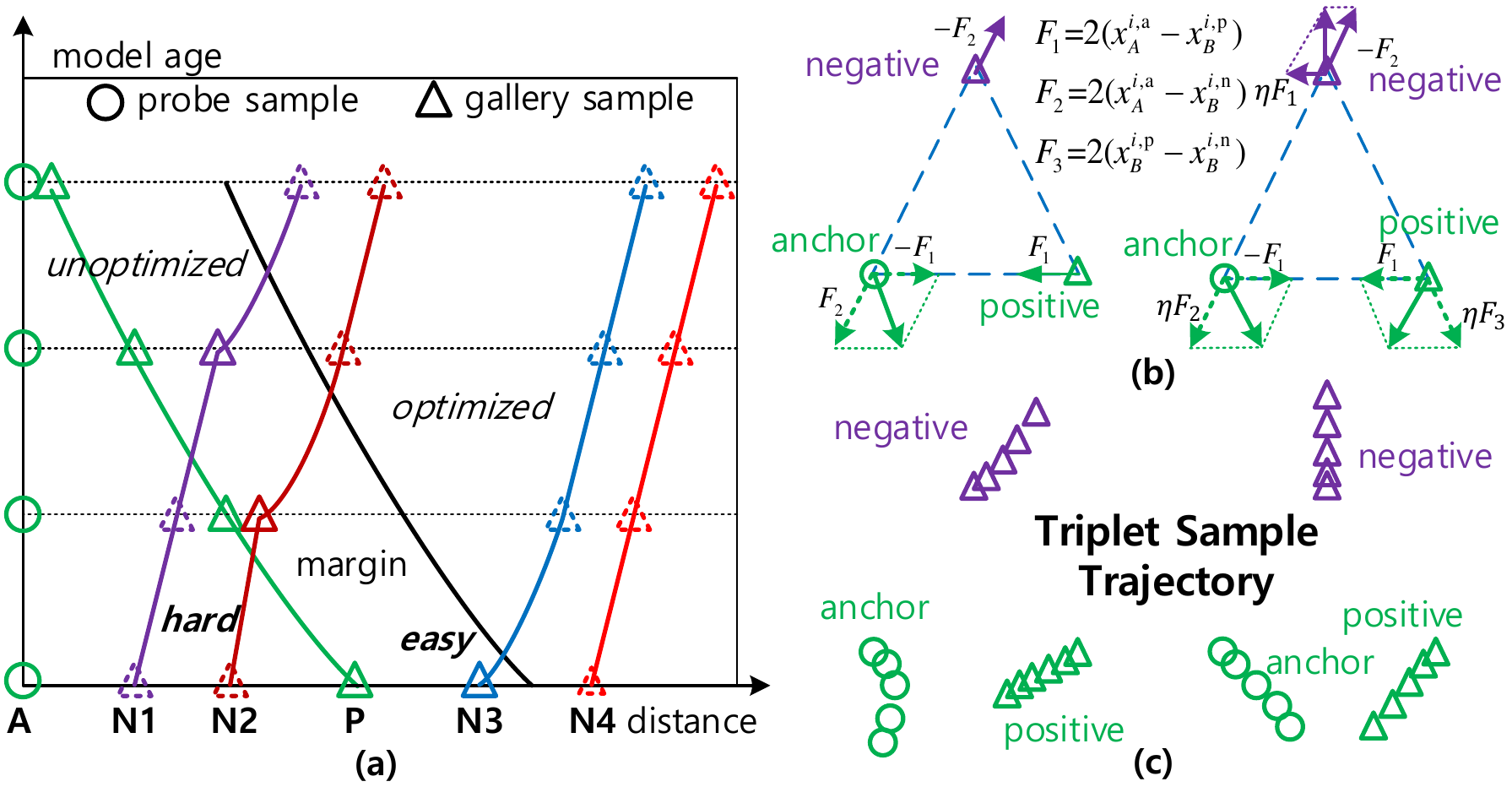}
    \end{tabular}
    \caption{Illustration of the self-paced learning strategy and symmetric gradient back-propagation constraint used in the training process. The left figure shows how the triplet units are gradually involved into the training process with the model age going on, in which the solid triangles denotes the current chosen training samples, and the dotted triangles represent the uninvolved or optimized training samples. The top right figure shows the gradient flow derived by the conventional triplet formulation and the one constrained by the symmetric regularizer term. The bottom right shows the two corresponding motion trajectories driven by the resulting gradient flows, in which our method can simultaneously minimize the intra-class distance and maximize the inter-class distance in each triplet.}
    \label{fig_3}
\end{figure}

The idea of our method is shown in Fig~\ref{fig_2}, in which we aim to learn a deep ranking model by using the relative similarity comparison in a self-paced manner. The deep ranking model learned in a self-paced manner can be formulated as follows:
\begin{equation}
\label{eq_2}
\mathrm{L} =  \hspace{-0.2cm}\sum \limits_{i=1}^N  \hspace{-0.1cm} u_i \mathcal{R}(\mathbf{x}_i^a, \hspace{-0.05cm}\mathbf{x}_i^p, \hspace{-0.05cm} \mathbf{x}_i^n)  +  \mathcal{G}(\mathbf{u}, \lambda, \vartheta) +  \zeta \mathcal{S}(\mathbf{x}_i^a, \hspace{-0.05cm}\mathbf{x}_i^p, \hspace{-0.05cm} \mathbf{x}_i^n)  +  \xi \mathcal{P}(\mathbf{W},\mathbf{b}),
\end{equation}
where $\mathbf{u} = [u_1, \dots, u_N]^T$ are the weights of all samples, $\lambda, \vartheta$ are the model age parameters, {\small $\zeta, \xi$} are the weights of regularizer term. Our method can jointly pull the positive pairs and push the negative pairs in each triplet unit. Specially, the relative similarity term $\mathcal{R}(\cdot)$ maximizes the relative distances between the positive pairs and negative pairs, the self-paced regularizer term $\mathcal{G}(\cdot)$ updates the sample weights in a self-paced manner, the symmetric regularizer term $\mathcal{S}(\cdot)$ revises the gradient back-propagation in a symmetric way, and the parameter regularizer term $\mathcal{P}(\cdot)$ smoothes the parameter of the deep CNN. In the following paragraphs, we explain these terms in detail.

{\bf Relative similarity term}  The relative similarity comparison metric has been widely applied in the object recognition communities, such as the face verification~\cite{Schroff_Kalenichenko_Philbin:2015} and person Re-ID~\cite{Ding_Lin_Wang:2015}, which is formulated as follows:
\begin{equation}
\label{eq_3}
\mathcal{R} = \max\{\mathcal{M} + \|f(\mathbf{x}_i^a) - f(\mathbf{x}_i^p)\|_2^2 - \|f(\mathbf{x}_i^a) - f(\mathbf{x}_i^n)\|_2^2, 0\},
\end{equation}
where $\mathcal{M}$ is the margin between positive pairs and negative pairs in the distance space, and $f(\cdot)$ is the learned feature mapping function. As a result, the relative distance between positive pairs and negative pairs are maximized, which is benefit to learn a deep ranking model in distinguishing the different individuals. As shown in Fig.~\ref{fig_3}, we argue that there are two drawbacks of directly applying this metric to solve the person Re-ID problem in deep learning framework, namely the equivalence of training samples and asymmetric gradient back-propagation, which will significantly weaken the generalization ability of the learned deep ranking model on the testing data.

\textbf{Self-paced regularizer term} In the SPL theory, a self-paced regularizer term is introduced to adaptively update the weights of samples according to both the training loss and model age. As shown in Fig.~\ref{fig_3} (a), the easy samples will contribute more than the hard samples when the model is young, and all the samples will be involved equally when the model is mature. For this purpose, we propose a novel soft polynomial regularizer term, which is formulated as follows:

\begin{equation}
\label{eq_4}
\mathcal{G} = \lambda (\frac{1}{t}\|\mathbf{u}\|_2^t - \frac{1}{\vartheta}\sum \limits_{i=1}^N u_i),
\end{equation}
where $\lambda > 0$ is the model age, $1 > \vartheta > 0$ is the mature age, and $t$ is the polynomial order. Different from the recent self-paced regularizers, such as hard weighting~\cite{Kumar_Packer_Koller:2010} and soft weighting~\cite{Zhao_Meng_Jiang:2015}, our method penalizes the loss according to the value of polynomial order. As a result, the weighting scheme deduced by our regularizer term can approach all of them.

\textbf{Symmetric regularizer term} The goal of our symmetric regularizer term is to revise the asymmetric gradient back-propagation deduced by the relative similarity comparison metric. As a result, the intra-class distance can be minimized and the inter-class distance can be maximized simultaneously in each triplet unit, as shown in Fig.~\ref{fig_3} (c). We penalize the deviation between two negative distances to keep the symmetric gradient back-propagation, which is formulated as follows:
\begin{equation}
\label{eq_5}
\mathcal{S} = \frac{1}{\gamma}\log \left(1 + \exp\left(\gamma \mathcal{Z}\right)\right),
\end{equation}
where $\mathcal{Z} = \left|\|f(\mathbf{x}_i^p) - f(\mathbf{x}_i^n)\|_2^2 - \|f(\mathbf{x}_i^a) - f(\mathbf{x}_i^n)\|_2^2\right|$ is the deviation measured in the Euclid distance, and $\gamma$ is the sharpness parameter. As shown in Fig.~\ref{fig_3} (b), we introduce $F_1$ and $F_3$ to jointly revise the back-propagation of negative sample and positive sample in each triplet unit. What's more~\footnote{For the detail analysis of how to control the direction, please refer to Eq.~\eqref{eq_10} in the optimization section.}, the strength and direction can be adaptively tuned according to the deviation.

\begin{figure*}[!htb]
\footnotesize
\centering
    \begin{tabular}{c}
        \includegraphics[height = 6.3cm, width = 17.2cm]{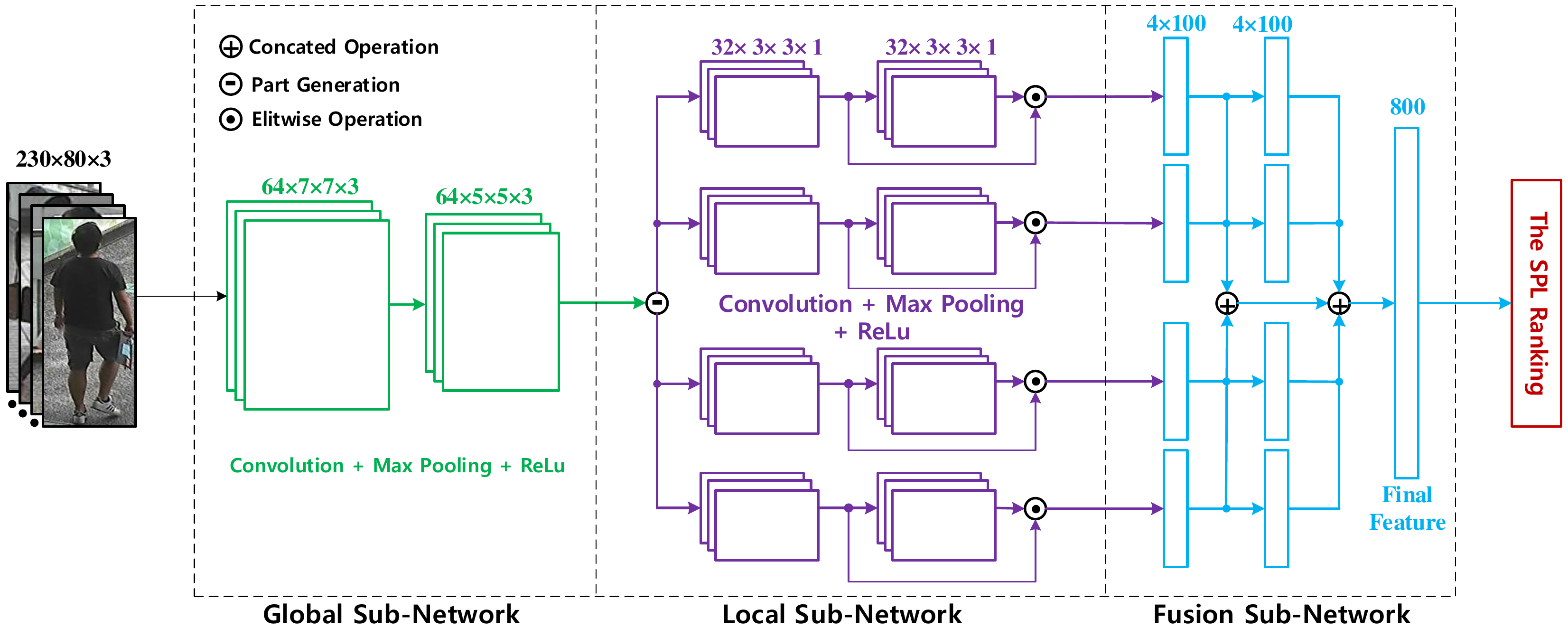}
    \end{tabular}
    \caption{The deep feature learning and fusion network. This architecture is comprised of three sub-networks: glob sub-network, local sub-network and fusion sub-network. The first two parts extract the global feature representations and local feature representations from person images by using convolutional layers, max-pooling layers and part generation strategy. The third parts learns and fuses the local feature representations from the second part by using fully connected layers. Finally, the concated feature representations are fed into the loss layer for similarity comparison.}
    \label{fig_4}
\end{figure*}

\textbf{Parameter regularizer term} In order to smooth the parameters of entire neural network, we define the following regularizer term:
\begin{equation}
\label{eq_6}
    \mathcal{P} = \sum\limits_{k = 1}^K \| \mathbf{W}^k\|_F^2 + \| \mathbf{b}^k\|_2^2\,,
\end{equation}
where $\|\cdot\|_F^2$ denotes the Frobenius norm, and $\|\cdot\|_2^2$ represents the Euclidian norm.

\subsection{Deep Neural Network}
In order to incorporate feature extraction and metric learning into an end-to-end framework, we propose a novel deep neural network which applies the part strategy to learn and fuse features from each individual. As shown in Fig.~\ref{fig_4}, the network is consisted of three subnetworks, which are introduced in the following paragraphs.

\textbf{Global subnetwork} It takes images in size of $230 \times 80 \times 3$ as input, and passes through two $64$ learned filters of size $7 \times 7 \times 3$ and $5 \times 5 \times 3$, respectively. Then, the resulting feature maps are passed through a max pooling kernel of size $3 \times 3$ with stride $3$. Finally, these feature maps are passed through a rectified linear unit~(ReLU).

\textbf{Local subnetwork} We firstly divided the input feature maps into four equal horizontal patches across the height channel, which introduces $4 \times 64$ local feature maps of different body parts. Then, we pass each local feature maps through two convolutional layers, and both of them have 32 learned filters of size $3 \times 3$. The outputs of the first and second local convolutional layer are summarized using eltwise operation. Afterwards, the resulting feature maps are passed through a max pooling kernel of size $3 \times 3$ with stride $1$. Finally, we add a rectified linear unit~(ReLU) after each max pooling layer.

\textbf{Fusion subnetwork} It takes local feature maps of different body parts as input, and learns discriminative features by concatenating two fully connected layers in each team. The dimension of the fully connected layers is {\small $100$} and a rectified linear unit~(ReLU) is added between them. Then, the resulting features of the first four fully connected layers are concatenated to be fused by adding another fully connected layer in dimension of {\small $400$}. Finally, the one {\small $400$} and four {\small $100$} dimensional features are concatenated to further generate the output {\small $800$} dimensional features.

\subsection{Optimization}
We use the gradient back-propagation method to optimize the parameters of deep CNN and weights of training samples. For simplicity, we consider the deep parameters as a whole and define $\mathbf{\Omega}^k = [\mathbf{W}^k, \mathbf{b}^k]$ and $\mathbf{\Omega} = \{\mathbf{\Omega}^1, \cdots, \mathbf{\Omega}^K\}$.

In order to employ the back-propagation algorithm to optimize the deep parameters, we compute the partial derivative of the loss function as follows:
\begin{equation}
\label{eq_7}
  \frac{\partial \mathrm{L}}{\partial \mathbf{\Omega}} =  \sum \limits_{i=1}^N  u_i r(\mathbf{x}_i^a, \mathbf{x}_i^p, \mathbf{x}_i^n) + \zeta s(\mathbf{x}_i^a, \mathbf{x}_i^p,  \mathbf{x}_i^n)  +  2 \xi \sum \limits_{k=1}^K \mathbf{\Omega}^k,
\end{equation}
where the three terms represent gradient of the relative similarity term, the symmetric regularizer term and the parameter regularizer term, respectively.

We define $\mathcal{T} = \mathcal{M} + \|f(\mathbf{x}_i^a) - f(\mathbf{x}_i^p)\|_2^2 - \|f(\mathbf{x}_i^a) - f(\mathbf{x}_i^n)\|_2^2$, therefore the gradient of relative similarity term can be formulated as follows:
\begin{equation}
\label{eq_8}
 r = \left\{
    \begin{array}{l}
        \frac{\partial \mathcal{R}(\mathbf{x}_i^a, \mathbf{x}_i^p, \mathbf{x}_i^n)}{\partial \mathbf{\Omega}}, \hspace{0.1cm} if \hspace{0.1cm} \mathcal{T} > 0;\\
        \hspace{0.8cm}0\hspace{0.85cm}, \hspace{0.1cm}else.
    \end{array} \right.,
\end{equation}
where $\frac{\partial \mathcal{R}}{\partial \mathbf{\Omega}}$ is formulated as follows:

\begin{equation}
\label{eq_9}
 \begin{aligned}
    \frac{\partial \mathcal{R}}{\partial \mathbf{\Omega}} =   2(f(\mathbf{x}_i^a) - f(\mathbf{x}_i^p))^{\prime} \cdot \frac{\partial f(\mathbf{x}_i^a)- \partial f(\mathbf{x}_i^p)}{\partial \mathbf{\Omega}}\\
     -2(f(\mathbf{x}_i^a) - f(\mathbf{x}_i^n))^{\prime} \cdot \frac{\partial f(\mathbf{x}_i^a)- \partial f(\mathbf{x}_i^n)}{\partial \mathbf{\Omega}}\\
     -2(f(\mathbf{x}_i^p) - f(\mathbf{x}_i^n))^{\prime} \cdot \frac{\partial f(\mathbf{x}_i^p)- \partial f(\mathbf{x}_i^n)}{\partial \mathbf{\Omega}}.
\end{aligned}
\end{equation}

\begin{figure}[!htb]
\footnotesize
\centering
    \begin{tabular}{cc}
        \includegraphics[height = 4.5cm, width = 4.3cm]{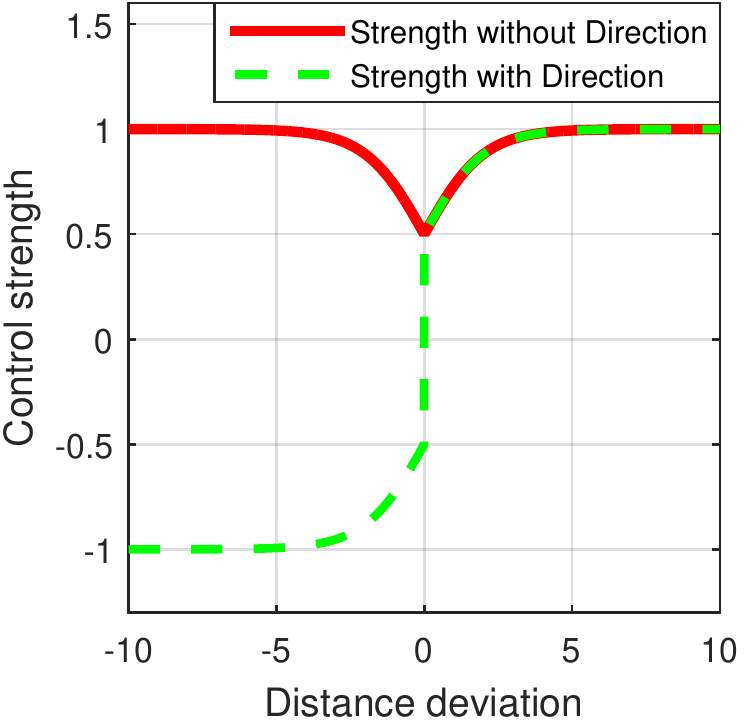} &
        \hspace{-0.5cm}
        \includegraphics[height = 4.5cm, width = 4.3cm]{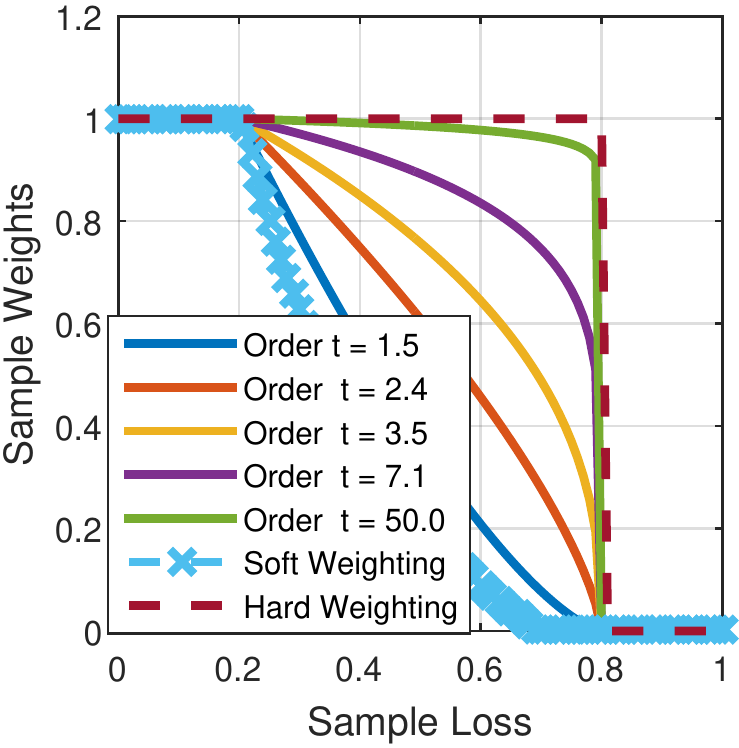}
    \end{tabular}
    \caption{Illustration of the symmetric back-propagation and comparison of different weighting schemes, in which the left one shows that the deduced strength and direction in controlling the gradient back-propagation can be adaptively updated according to the distance derivation, and the right one shows that our method can approach nearly all the weighting schemes by just tuning the polynomial order.}
    \label{fig_5}
\end{figure}

By defining $\mathcal{D}= \|f(\mathbf{x}_i^p) - f(\mathbf{x}_i^n)\|_2^2 - \|f(\mathbf{x}_i^a) - f(\mathbf{x}_i^n)\|_2^2$, then the gradient of symmetric regularizer term can be formulated as follows:
\begin{equation}
\label{eq_10}
 s = \eta \hspace{0.05cm} \mathrm{sign}(\mathcal{D}) \cdot \frac{\partial \mathcal{D}(\mathbf{x}_i^a, \mathbf{x}_i^p, \mathbf{x}_i^n)}{\partial \mathbf{\Omega}},
\end{equation}
where {\small $\eta = \exp(\gamma \mathcal{Z})/(1 +\exp(\gamma \mathcal{Z}))$} and $\mathrm{sign}(\mathcal{D})$ denote the strength and direction in the symmetric back-propagation, and $\frac{\partial \mathcal{D}}{\partial \mathbf{\Omega}}$ is formulated as follows:
\begin{equation}
\label{eq_11}
\begin{aligned}
   \frac{\partial \mathcal{D}}{\partial \mathbf{\Omega}}  =  2 (f(\mathbf{x}_i^a) - f(\mathbf{x}_i^n))^{\prime} \cdot \frac{\partial f(\mathbf{x}_i^a)- \partial f(\mathbf{x}_i^n)}{\partial \mathbf{\Omega}}\\
     -2 (f(\mathbf{x}_i^p) - f(\mathbf{x}_i^n))^{\prime} \cdot \frac{\partial f(\mathbf{x}_i^p)- \partial f(\mathbf{x}_i^n)}{\partial \mathbf{\Omega}}
\end{aligned}.
\end{equation}
As shown in Fig.~\ref{fig_5}, the deduced strength and direction in controlling the gradient back-propagation can be adaptively updated according to the distance derivation, which is benefit to promote the symmetric back-propagation.

In order to update the weights of samples in each iteration, we deduce the closed form solution of the SPL model under the proposed regularizer term. Because the soft polynomial regularizer is convex in $[0,1]$, it is easy to derive the optimal solution to $\min_{\mathbf{u}\in [0,1]} \sum\nolimits_{i = 1}^N {u_i} \mathcal{R} + \mathcal{G}(\mathbf{u}, \lambda, \vartheta)$ as follows:

\begin{equation}
\label{eq_12}
 u_i^{*} = \left\{
    \begin{array}{l}
        1, \hspace{2.5cm} if \hspace{0.1cm} \mathcal{R} < \lambda(\frac{1}{\vartheta} - 1),\\
        0, \hspace{2.5cm} if \hspace{0.1cm} \mathcal{R} > \frac{\lambda}{\vartheta},\\
        (\frac{1}{\vartheta} - \frac{R}{\lambda})^{1/(t-1)}, \hspace{0.1cm} otherwise.
    \end{array} \right..
\end{equation}
The comparison with hard and soft weighting schemes are shown in Fig.~\ref{fig_5}, in which our method can approach them by tuning the polynomial order. If the loss is smaller than a threshold ${\lambda}/{\vartheta}$, it will be treated as an easy sample and assigned a positive weight; If the loss is further smaller ${\lambda}(1/{\vartheta} - 1)$, the sample is treated as a faithful sample weighted by $1$. Therefore, the easy-sample-first property~\cite{Kumar_Packer_Koller:2010} and soft weighting strategy~\cite{Zhao_Meng_Jiang:2015} are all inherited in our method.

From the above derivations, we can see that the deep parameters and sample weights can be easily optimized given the values of {\small $f(\mathbf{x}_i^a)$}, {\small $f(\mathbf{x}_i^p)$}, {\small $f(\mathbf{x}_i^n)$} and {\small $\partial f(\mathbf{x}_i^a)/\partial \mathbf{\Omega}$}, {\small $\partial f(\mathbf{x}_i^p)/\partial \mathbf{\Omega}$}, {\small $\partial f(\mathbf{x}_i^n)/\partial \mathbf{\Omega}$}, in which they can be obtained by separately running the forward and backward propagation by traveling all the triplets in each mini-batch. As the algorithm needs to accumulate the gradients in a self-paced way, we call it the self-paced gradient descent algorithm. We show the process in Algorithm~\ref{alg}.

\begin{algorithm}[tb]
   \caption{The self-paced gradient descent algorithm}
   \label{alg}
\begin{algorithmic}
   \State {\bfseries Input:}\\
    \hspace{0.5cm} The training triplets {\small $\mathbf{X}$}, learning rate $\tau$, age updating rate $\omega$, maximum iterations $H$, margin parameters $\mathcal{M}$, age parameters $\lambda, \vartheta$, weight parameters $\zeta, \xi$ and sharpness parameter $\gamma$.
   \State {\bfseries Output:}\\
   \hspace{0.5cm} The network parameters $\mathbf{\Omega}$.
   \State {\bfseries repeat}
   \State \hspace{0.5cm} 1. Given an anchor sample $\mathbf{x}_i^a$, we randomly generate 200 triplets for each anchor in a mini-batch. Then, we calculate the output features of {\small $f(\mathbf{x}_i^a)$}, {\small $f(\mathbf{x}_i^p)$} and {\small $f(\mathbf{x}_i^n)$} of all the triplets by forward propagation.
   \State \hspace{0.5cm} {\bfseries repeat}
   \State \hspace{1.0cm} a) Update the sample weights parameters $\mathbf{u}$ according to Eq.~\eqref{eq_3} and Eq.~\eqref{eq_12};
   \State \hspace{1.0cm} b) Calculate {\small $\frac{\partial \mathcal{R}}{\partial \mathbf{\Omega }}$},{\small $\frac{\partial \mathcal{D}}{\partial \mathbf{\Omega }}$} according to Eq.~\eqref{eq_9} and Eq.~\eqref{eq_11};
   \State \hspace{1.0cm} c) Increment the gradient {\small $\frac{\partial \mathrm{L}}{\partial \mathbf{\Omega }}$} according to Eq.~\eqref{eq_7};
   \State \hspace{0.5cm} {\bfseries until} Travel all the triplet units in each mini-batch.
   \State \hspace{0.5cm} 2. Update the deep parameter $\mathbf{\Omega}_{h+1} = \mathbf{\Omega}_h - {\tau_h}\frac{\partial \mathrm{L}}{\partial \mathbf{\Omega}_h}$ and model age $\lambda_{h+1} = \lambda_h/{\omega_h}$, with $h \leftarrow h + 1$.
   \State {\bfseries until} $h > H$
\end{algorithmic}
\end{algorithm}

\section{Experiment}
\label{sec_experiment}
In this section, we firstly introduce the datasets, the parameter setting and the evaluation protocol. Then, we evaluate the performance of our approach on five benchmark datasets, respectively. Finally, we give a detailed analysis of the experimental results.
\subsection{Datasets and Settings}
{\bf Datasets:} Our experiments were conducted on five public datasets: the VIPeR~\cite{Gray_Tao:2008}, 3DPeS~\cite{Baltieri_Vezzani_Cucchiara:2011}, CUHK01~\cite{Li_Wang:2013}, CUHK03~\cite{Li_Zhao_Xiao:2014} and Market1501~\cite{Zheng_Shen_Tian:2015}. Specially, VIPeR has 632 pedestrian image pairs captured outdoor with varying viewpoints and illumination conditions. 3DPeS contains 1011 image of 192 pedestrians captured from 8 outdoor camera views with significantly different viewpoints. CHUK01 contains 971 pedestrians from two disjoint camera views. CUHK03 has 13164 images of 1360 pedestrians captured by six different cameras. Market1501 contains 32668 images of 1501 identities, in which each identity is captured by six cameras at most and two cameras at least. Each pedestrian has at least one sample under each camera view.

{\bf Parameter setting:} The weights are initialized from two zero-mean Gaussian distribution with the standard deviations from $0.01$ to $0.001$, respectively. The bias terms are set to $0$. The learning rate $\tau = 0.01$, age updating rate $\omega = 0.9$, weight parameters $\zeta = 0.1, \xi = 0.01$, sharpness parameter $\gamma = 0.9$, age parameters $\lambda = 0.6, \vartheta = 0.75$ and margin parameter $\mathcal{M} = 1.1$.

{\bf Evaluation protocol:} Our experiment follows the single-shot protocol in~\cite{Ding_Lin_Wang:2015}, in which 316 pedestrians in the VIPeR dataset, 96 pedestrians from the 3DPeS dataset, and 871 pedestrians of the CUHK01 dataset are randomly chosen to train the network, and the others are used to evaluate the performance. For the CUHK03 and Market1501 datasets, we use the provided fixed training and testing set in our experiment. The performance on the CUHK03 dataset is evaluated under the same single-shot protocol, and the performance on the Market1501 dataset is evaluated under both the single-query and multi-query evaluation settings as in~\cite{Zhang_Xiang_Gong:2016}. The cumulative matching characteristic~(CMC) curve is used to measure the performance of each method, which is an estimation of finding the corrected top $n$ match. Besides, the mAP is also used to evaluate the performance on the Market1501 dataset. To obtain the statistical results, we repeated the testing $10$ times and reported the average results.

\subsection{Results}

\begin{table}[h]
\footnotesize
\caption{Matching rates(\%) on the VIPeR dataset.}
\begin{center}
\label{tab_1}
\begin{tabular}{ c| c | c | c| c | c}
\hline
Methods & Top1 & Top5 & Top10 & Top15 & Top20\\
\hline
\hline
LOMO+XQDA~\cite{Liao_Hu_Zhu:2015}          & 40.00 & 68.13 & 80.51 & 87.37 & 91.08 \\
Quadruplet~\cite{Chen_Chen_Zhang:2017}     & 49.05 & 73.10 & 81.96 & $--$ & $--$ \\
ME~\cite{Paisitkriangkrai_Shen_Van:2015}   & 45.89 & 77.40 & 88.87 & 93.52 & 95.84 \\
LMF+LADF~\cite{Zhao_Ouyang_Wang:2014}      & 43.39 & 73.04 & 84.87 & 90.85 & 93.70 \\
SCSP~\cite{Chen_Yuan_Chen:2016}            & \textcolor{blue}{53.54} & \textcolor{blue}{82.59} & \textcolor{blue}{91.49} & \textcolor{red}{\bf{95.09}} & \textcolor{red}{\bf{96.65}} \\
\hline
Our method (Baseline)       & 45.57 & 68.67 & 78.48 & 81.65 & 83.86 \\
Our method (DSPL)           & \textcolor{red}{\bf{56.32}}  & \textcolor{red}{\bf{83.04}}  & \textcolor{red}{\bf{92.01}}
                   & \textcolor{blue}{93.78}  & \textcolor{blue}{95.88} \\
\hline
\end{tabular}
\end{center}
\end{table}

\begin{table}[h]
\footnotesize
\caption{Matching rates(\%) on the 3DPeS dataset.}
\begin{center}
\label{tab_2}
\begin{tabular}{ c| c | c | c| c | c}
\hline
Methods & Top1 & Top5 & Top10 & Top15 & Top20\\
\hline
\hline
KISSME~\cite{Koestinger_Hirzer_Wohlhart:2012} & 22.94 & 48.71 & 62.21 & 72.39 & 78.11 \\
LF~\cite{Pedagadi_Orwell_Velastin:2013}       & 33.43 & 45.50 & 69.98 & 76.53 & 81.03 \\
ME~\cite{Paisitkriangkrai_Shen_Van:2015}          & 53.30 & 76.79 & 86.03 & 89.37 & 92.78 \\
kLFDA~\cite{Xiong_Gou_Camps:2014}             & 54.02 & 77.74 & 85.92 & 90.04 & 92.38 \\
SCSP~\cite{Chen_Yuan_Chen:2016}               & 57.29 & 78.97 & 85.01 & 89.52 & 91.51 \\
\hline
Our Method (Baseline)                         &\textcolor{blue}{63.38} &\textcolor{blue}{85.42} &\textcolor{blue}{93.21}
                                              &\textcolor{blue}{93.78} &\textcolor{blue}{95.07} \\
Our Method (DSPL)                         &\textcolor{red}{\bf{72.23}}  &\textcolor{red}{\bf{90.69}}  &\textcolor{red}{\bf{95.34}}
                                              &\textcolor{red}{\bf{96.78}} &\textcolor{red}{\bf{97.51}}  \\
\hline
\end{tabular}
\end{center}
\end{table}

We compare our results with the following methods, namely the Quadruplet~\cite{Chen_Chen_Zhang:2017}, Bow~\cite{Zheng_Shen_Tian:2015}, kLFDA~\cite{Xiong_Gou_Camps:2014}, SCSP~\cite{Chen_Yuan_Chen:2016}, FPNN~\cite{Li_Zhao_Xiao:2014}, LDNS~\cite{Zhang_Xiang_Gong:2016}, JSC~\cite{Wang_Zuo_Lin:2016}, LMNN~\cite{Weinberger_Saul:2009}, LOMO+XQDA~\cite{Liao_Hu_Zhu:2015}, IDLA~\cite{Ahmed_Jones_Marks:2015}, CDVM~\cite{Lin_Wang_Zuo:2016},  LMF+LADF~\cite{Zhao_Ouyang_Wang:2014}, KISSME~\cite{Koestinger_Hirzer_Wohlhart:2012}, LSSCDL~\cite{Zhang_Li_Lu:2016} and ME~\cite{Paisitkriangkrai_Shen_Van:2015}. In order to show how much our DSPL method contributes to the performance, we take the results of relative similarity comparison metric as baseline. Comparison results on the five datasets are shown in Table~\ref{tab_1} to Table~\ref{tab_5}, respectively. In these tables, the best performance is highlighted in red, and the second best is is highlighted in blue.

For the VIPeR dataset, we compare our method with both the traditional methods and the deep learning based method, as shown in Table~\ref{tab_1}. From the results, we can see that our DSPL outperforms the baseline method with $10.75\%$ in Top 1 accuracy, which demonstrates its effective by introducing the SPL and symmetric gradient back-propagation regularization. What's more, it outperforms the previous best performed method SCSP~\cite{Chen_Yuan_Chen:2016} with $2.78\%$ in Top 1 accuracy.

\begin{table}[h]
\footnotesize
\caption{Matching rates(\%) on the CUHK01 dataset.}
\begin{center}
\label{tab_3}
\begin{tabular}{ c| c | c | c| c | c}
\hline
Methods & Top1 & Top5 & Top10 & Top15 & Top20\\
\hline
\hline
KISSME~\cite{Koestinger_Hirzer_Wohlhart:2012}  & 29.40 & 59.34 & 71.45 & 80.09 & 88.12 \\
LMNN~\cite{Weinberger_Saul:2009}               & 21.17 & 49.49 & 61.12 & 69.93 & 78.32 \\
IDLA~\cite{Ahmed_Jones_Marks:2015}             & 65.00 & 89.33 & 92.04 & 93.74 & 96.51 \\
JSC~\cite{Wang_Zuo_Lin:2016}                   & 65.71 & 89.41 & 92.52 & 93.74 & 96.63 \\
CDVM~\cite{Lin_Wang_Zuo:2016}                  & 66.50 & \textcolor{blue}{93.00} & \textcolor{blue}{96.50} & \textcolor{blue}{99.00} & \textcolor{blue}{99.00} \\
\hline
Our method (Baseline)         & \textcolor{blue}{71.14} & 89.12 & 94.19 & 97.85 & 98.84 \\
Our method (DSPL)             &\textcolor{red}{\bf{81.33}}  &\textcolor{red}{\bf{94.35}}  &\textcolor{red}{\bf{98.23}}
                                              &\textcolor{red}{\bf{100.00}} &\textcolor{red}{\bf{100.00}}  \\
\hline
\end{tabular}
\end{center}
\end{table}

Table~\ref{tab_2} lists the results on the 3DPeS dataset, in which our baseline method gets the second best performance, contributed by the part-based deep CNN architecture, and our DSPL method achieves the best performance in all Top 1 to Top 20 accuracies. Compared with previous best performed method SCSP~\cite{Chen_Yuan_Chen:2016} on this dataset, our two methods outperform it by $6.09\%$ and $14.94\%$ in Top 1 accuracy, respectively. In addition, benefit from the SPL strategy and symmetric gradient back-propagation constraint used in our method, the DSPL method wins the baseline method $8.85\%$ in Top 1 accuracy.

In Table~\ref{tab_3}, we report the comparison results with the state-of-the-art methods on the CUHK01 dataset, in which our DSPL method achieves the best performance in all comparison groups from Top 1 to Top 20. Specially, our baseline method outperforms the previous best method CDVM~\cite{Lin_Wang_Zuo:2016} with $4.64\%$ in Top 1, which demonstrates the effective by applying the part-based CNN to extract the feature representations. What's more, our DSPL method outperforms the baseline method and CDVM method with $10.19\%$ and $14.83\%$ in Top 1, respectively.

For the CUHK03 dataset, we compare our method with several state-of-the-art methods. The detail results are shown in Table~\ref{tab_4}, in which our DSPL method also achieves the best performance in all comparison groups from Top 1 to Top 20. Compared with the previous best method CDVM~\cite{Lin_Wang_Zuo:2016}, our baseline method fall behind it by $1.08\%$  in Top 1 accuracy, while our DSPL method outperforms it by $14.77\%$ in Top 1 accuracy. Benefit from the DSPL and symmetric regularizer used in our method, the DSPL method wins the baseline method by $15.85\%$ in Top 1 accuracy.

\begin{figure*}[!htb]
\footnotesize
\centering
    \begin{tabular}{ccccc}
        \hspace{-0.5cm}
        \includegraphics[height = 3.7cm, width = 3.7cm]{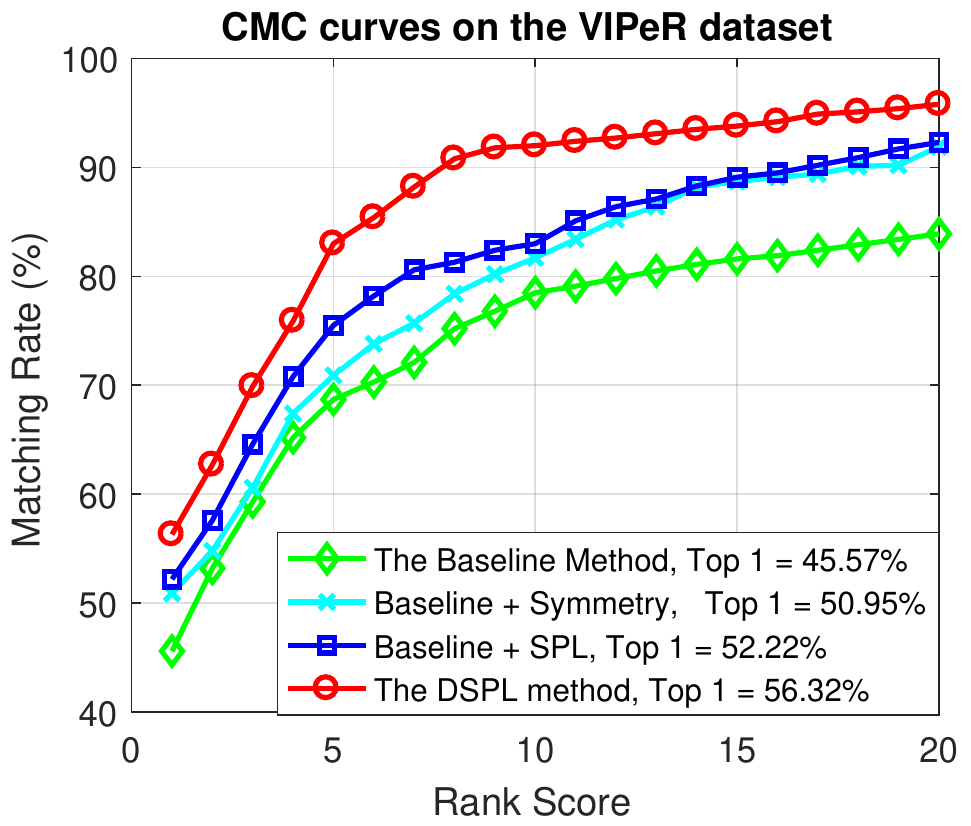} &
        \hspace{-0.5cm}
        \includegraphics[height = 3.7cm, width = 3.7cm]{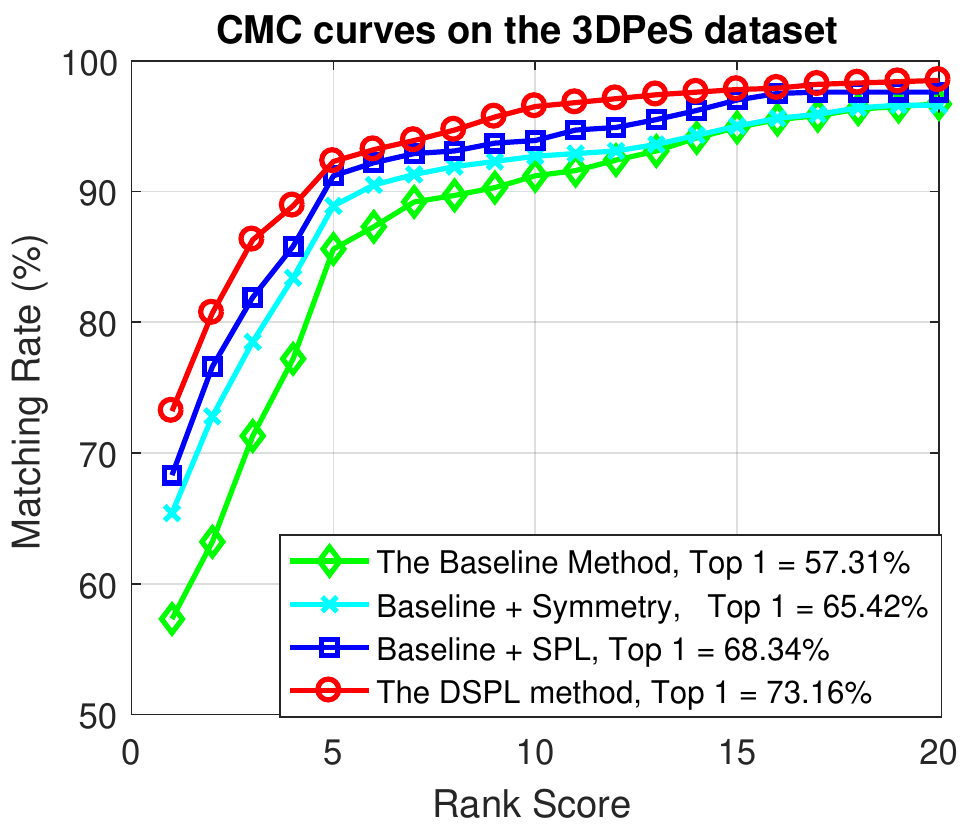} &
        \hspace{-0.5cm}
        \includegraphics[height = 3.7cm, width = 3.7cm]{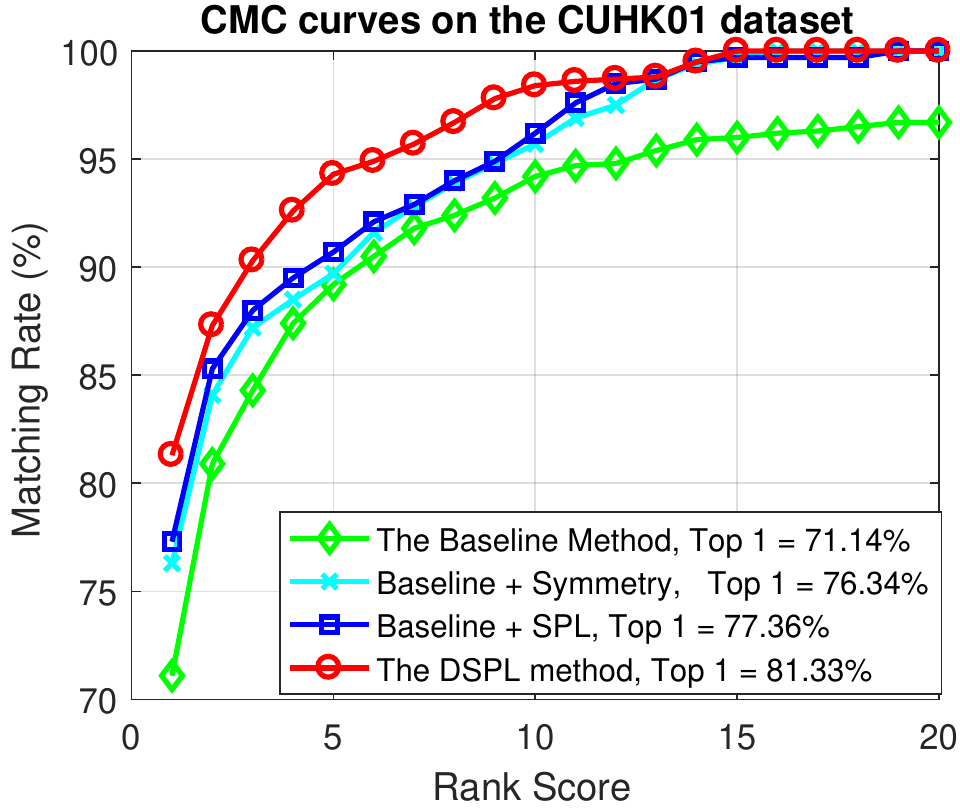}&
        \hspace{-0.5cm}
        \includegraphics[height = 3.7cm, width = 3.7cm]{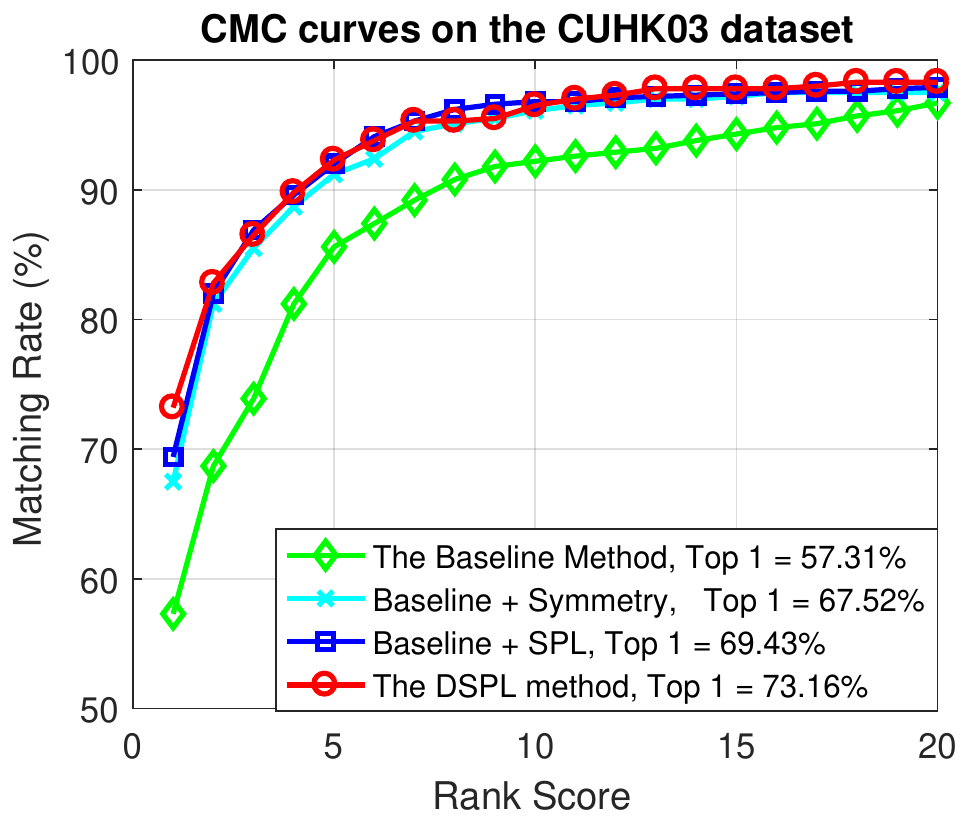}&
        \hspace{-0.5cm}
        \includegraphics[height = 3.7cm, width = 3.7cm]{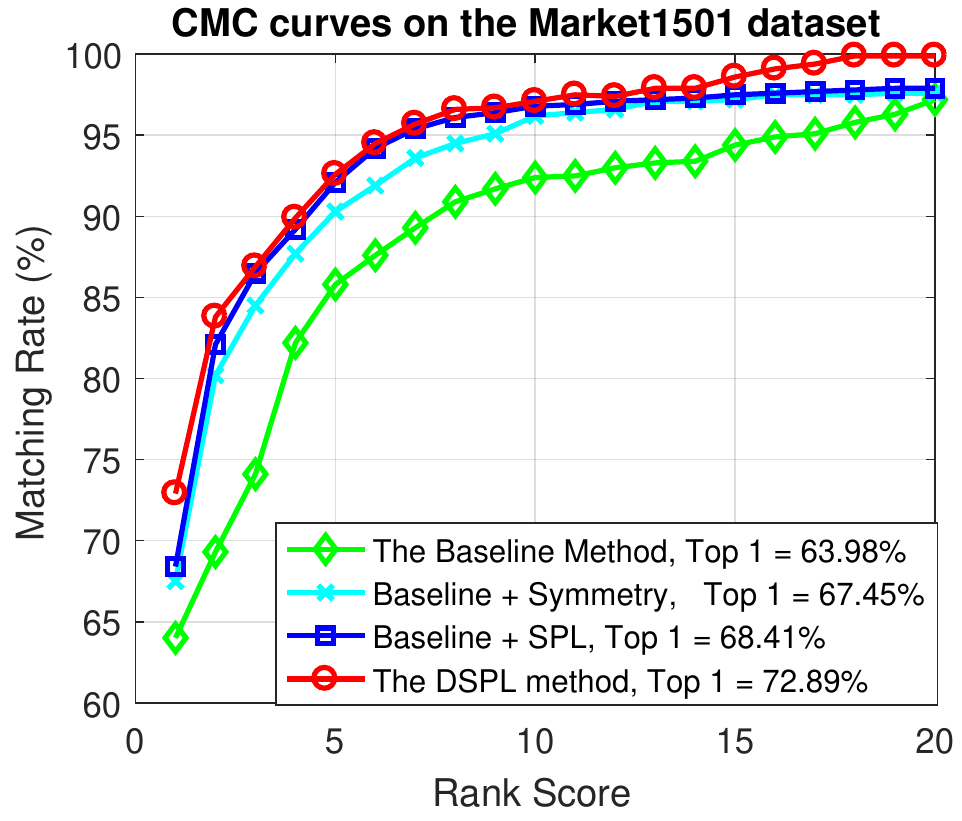}
    \end{tabular}
    \caption{Comparison results of each component that contributes to the final performance. From these results, we can conclude that the SPL training strategy and symmetric constraint can continuously improve the identification performance and our DSPL method can get the best performance by incorporating the two constraints into an end-to-end learning framework.}
    \label{fig_6}
\end{figure*}

\begin{table}[h]
\footnotesize
\caption{Matching rates(\%) on the CUHK03 dataset.}
\begin{center}
\label{tab_4}
\begin{tabular}{ c| c | c | c| c | c}
\hline
Methods & Top1 & Top5 & Top10 & Top15 & Top20\\
\hline
\hline
LOMO+XQDA~\cite{Liao_Hu_Zhu:2015}             & 52.20 & 81.29 & 90.94 & 94.21 & 95.01 \\
FPNN~\cite{Li_Zhao_Xiao:2014}                 & 20.65 & 51.02 & 68.83 & 76.38 & 81.45 \\
IDLA~\cite{Ahmed_Jones_Marks:2015}            & 54.74 & 87.59 & 94.01 & 95.02 & 95.41\\
LSSCDL~\cite{Zhang_Li_Lu:2016}                & 57.00 & 84.38 & 90.93 & 94.32 & 95.12\\
CDVM~\cite{Lin_Wang_Zuo:2016}                 & \textcolor{blue}{58.39} & 85.56 & \textcolor{blue}{92.57} & 94.48 & 96.60\\
\hline
Our method (Baseline)         & 57.31 & \textcolor{blue}{85.62} & 92.19 & \textcolor{blue}{94.85} & \textcolor{blue}{96.74} \\
Our method (DSPL)             & \textcolor{red}{\bf{73.16}} & \textcolor{red}{\bf{92.26}}  & \textcolor{red}{\bf{96.54}}
                     & \textcolor{red}{\bf{97.75}} & \textcolor{red}{\bf{98.45}}  \\
\hline
\end{tabular}
\end{center}
\end{table}

\begin{table}[h]
\footnotesize
\caption{Matching rates(\%) on the Market1501 dataset.}
\begin{center}
\label{tab_5}
\begin{tabular}{ c| c | c || c | c  c}
\hline
\multicolumn{1}{c|}{\multirow{2}{*}{Methods}} &
\multicolumn{2}{c||}{Single-Query} &
\multicolumn{2}{c}{Multi-Query}\\
\cline{2-5}
& Top1 & mAP & Top1  & mAP\\
\hline
\hline
Bow~\cite{Zheng_Shen_Tian:2015}                         & 34.38  & 14.10 & 42.64 & 19.47\\
kLFDA~\cite{Xiong_Gou_Camps:2014}                       & 51.37  & 24.43 & 52.67 & 27.36\\
KISSME~\cite{Koestinger_Hirzer_Wohlhart:2012}           & 40.50  & 19.02 & $--$  & $--$ \\
LDNS~\cite{Zhang_Xiang_Gong:2016}                       & 61.02  & 35.68 & 71.56 & 46.03 \\
SCSP~\cite{Chen_Yuan_Chen:2016}                         & 51.90  & 26.35 & $--$  & $--$\\
\hline
Our Method (Baseline)                                   &\textcolor{blue}{\bf{63.98}} & \textcolor{blue}{\bf{38.21}}
                                                        &\textcolor{blue}{\bf{79.52}} & \textcolor{blue}{\bf{53.01}}\\
Our Method (DSPL)                                       &\textcolor{red}{\bf{72.89}}  & \textcolor{red}{\bf{46.68}}
                                                        &\textcolor{red}{\bf{87.05}}  & \textcolor{red}{\bf{57.41}}\\
\hline
\end{tabular}
\end{center}
\end{table}

Finally, the Market1501 dataset is a newly proposed large scale dataset for person Re-ID. The best performance was obtained by a conventional method LDNS~\cite{Zhang_Xiang_Gong:2016}. As illustrated in Table~\ref{tab_5}, the proposed two methods outperform LDNS by $2.96\%$ and $11.87\%$ in Top 1 accuracy under the single-query setting, and $7.96\%$ and $15.49\%$ in Top 1 accuracy under the multi-query setting, respectively. Again, our DSPL method wins the baseline method by $8.91\%$ and $7.53\%$ in Top 1 accuracy under the single-query and the multi-query evaluation settings, respectively. For the mAP evaluation, the same conclusion can be made.

\begin{table}[h]
\footnotesize
\caption{Matching rates by using different networks.}
\begin{center}
\label{tab_6}
\begin{tabular}{ c| c | c | c| c | c}
\hline
Networks & VIPeR & 3DPeS & CUHK01 & CUHK03 & Market501\\
\hline
\hline
G-Net       & 39.28 & 50.31 & 54.35 & 42.12 & 45.54\\
P-Net         & \textcolor{red}{\bf{56.32}} & \textcolor{red}{\bf{72.23}} & \textcolor{red}{\bf{81.33}}
                    & \textcolor{red}{\bf{73.16}} & \textcolor{red}{\bf{72.89}}\\
\hline
\end{tabular}
\end{center}
\end{table}

\subsection{Analysis}
In this section, we analyze the experimental results of our method from two aspects, namely the effectiveness of each contribution to the final performance and the robustness of our method to different parameter settings. The detailed analysis results are given in the following paragraphs.

{\bf Effectiveness of each contribution}  Firstly, we report some intermediate results of each contribution in our method, namely the baseline method, the baseline method + symmetric regularizer, the baseline method + SPL and the DSPL method, so as to illustrate how much each of them contributes to the ranking performance on the five datasets, respectively. The comparison results are shown in Fig.~\ref{fig_6}, from which we can make the following two conclusions: 1) The baseline method can be improved by separately incorporating the SPL strategy or the symmetric regularizer into the relative similarity comparison framework; 2) The DSPL method significantly outperforms the baseline method and the other two methods by jointly introducing the SPL strategy and symmetric regularizer into an end-to-end learning framework. The above two points tell us that it is very important to distinguish the reliability of samples and revise the gradient back-propagation in the training process.

\begin{figure*}[!htb]
\footnotesize
\centering
    \begin{tabular}{ccccc}
        \hspace{-0.5cm}
        \includegraphics[height = 3.9cm, width = 3.6cm]{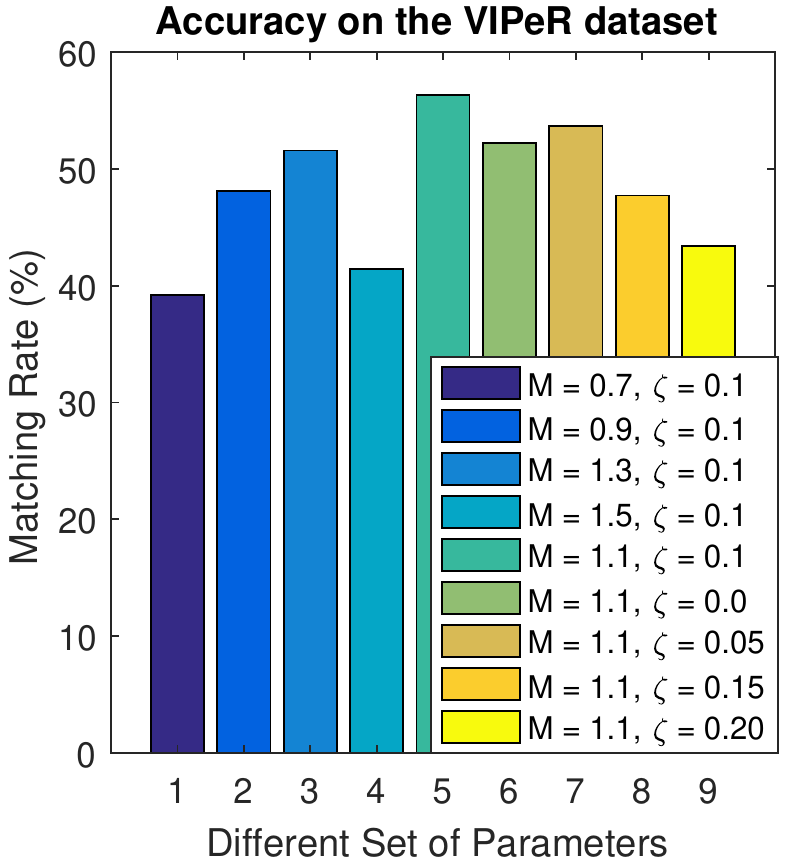} &
        \hspace{-0.5cm}
        \includegraphics[height = 3.9cm, width = 3.6cm]{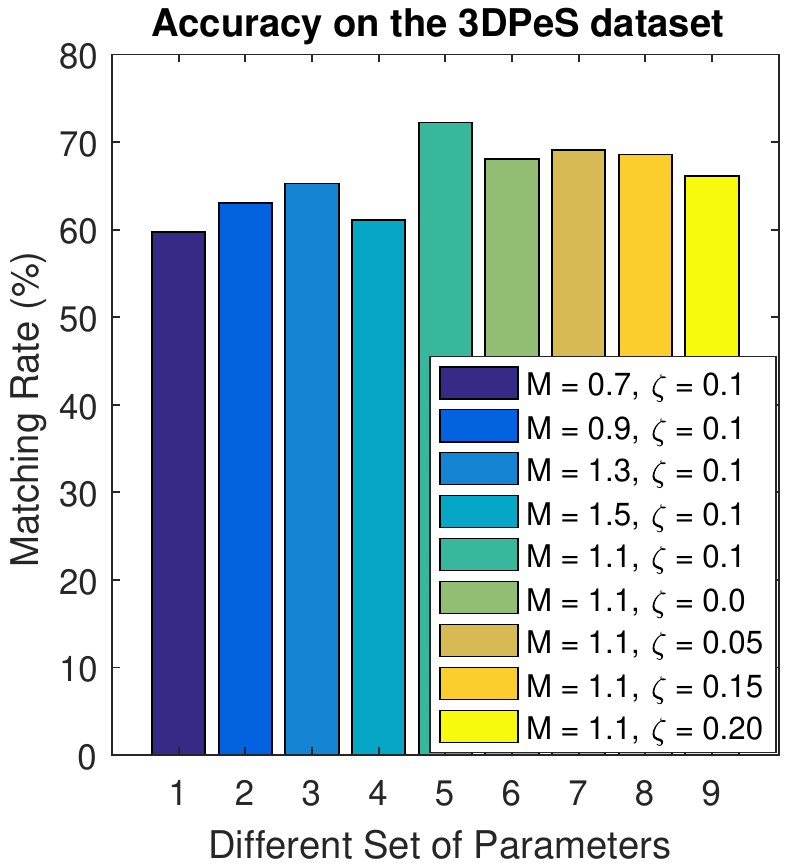} &
        \hspace{-0.5cm}
        \includegraphics[height = 3.9cm, width = 3.6cm]{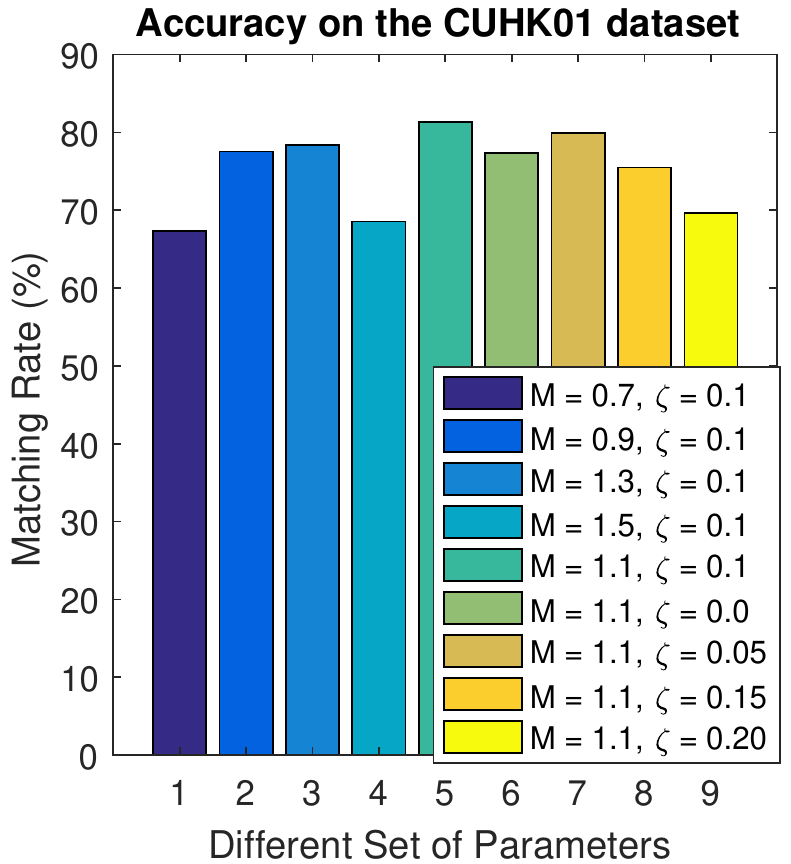}&
        \hspace{-0.5cm}
        \includegraphics[height = 3.9cm, width = 3.6cm]{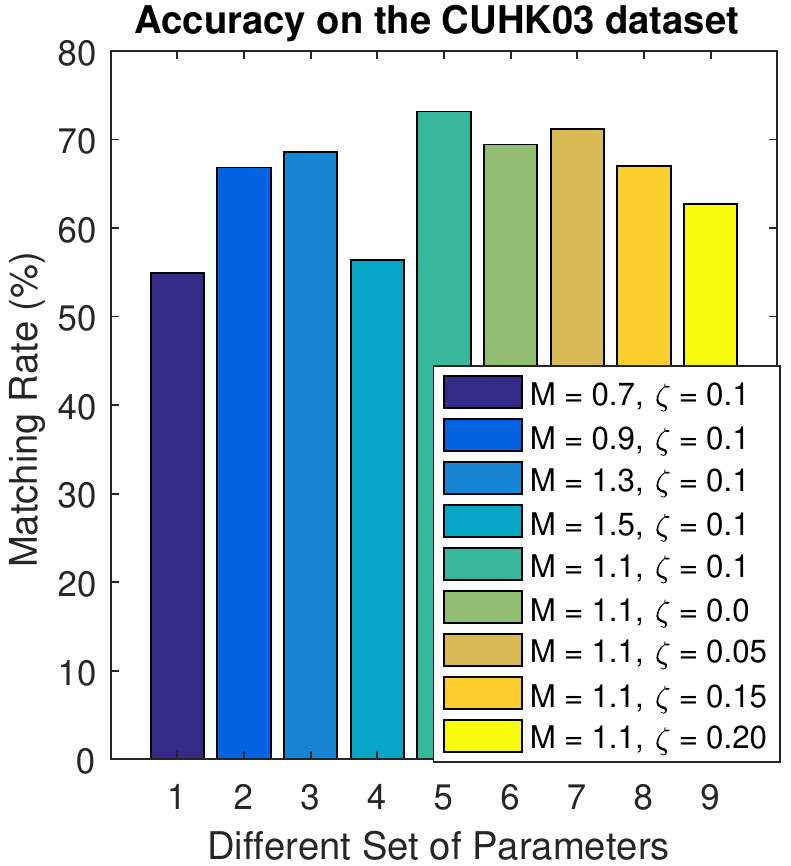}&
        \hspace{-0.5cm}
        \includegraphics[height = 3.9cm, width = 3.6cm]{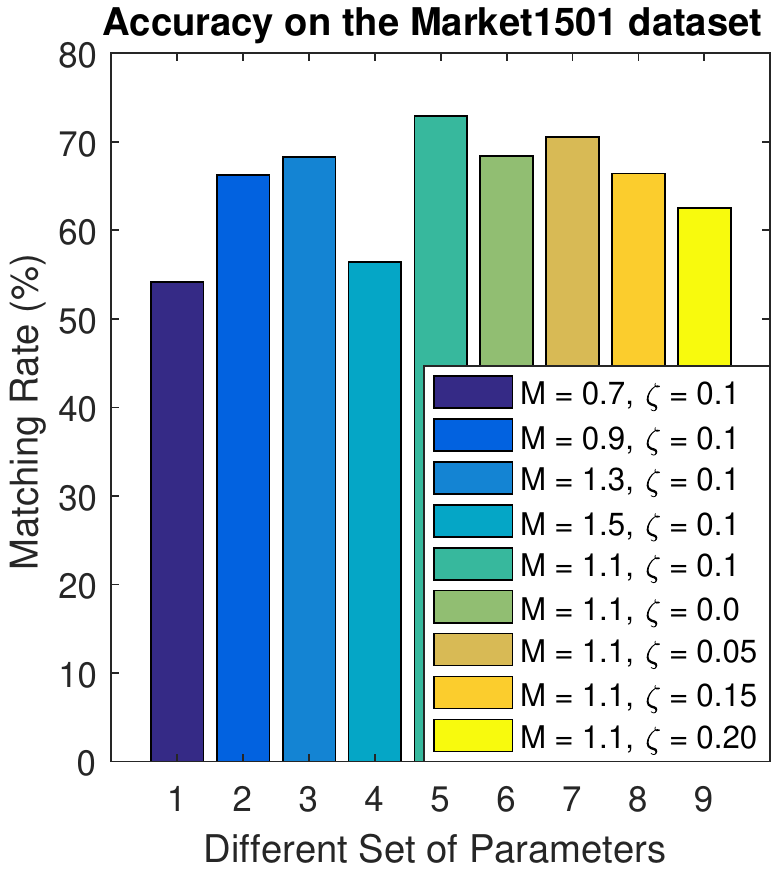}\\
        \hspace{-0.5cm}
        \includegraphics[height = 3.9cm, width = 3.6cm]{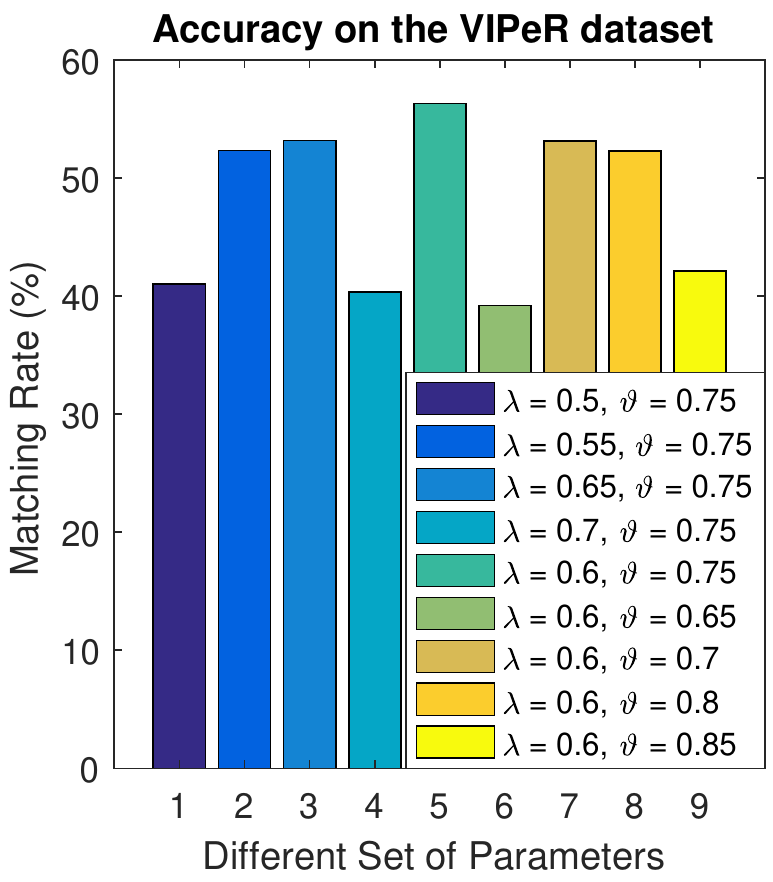} &
        \hspace{-0.5cm}
        \includegraphics[height = 3.9cm, width = 3.6cm]{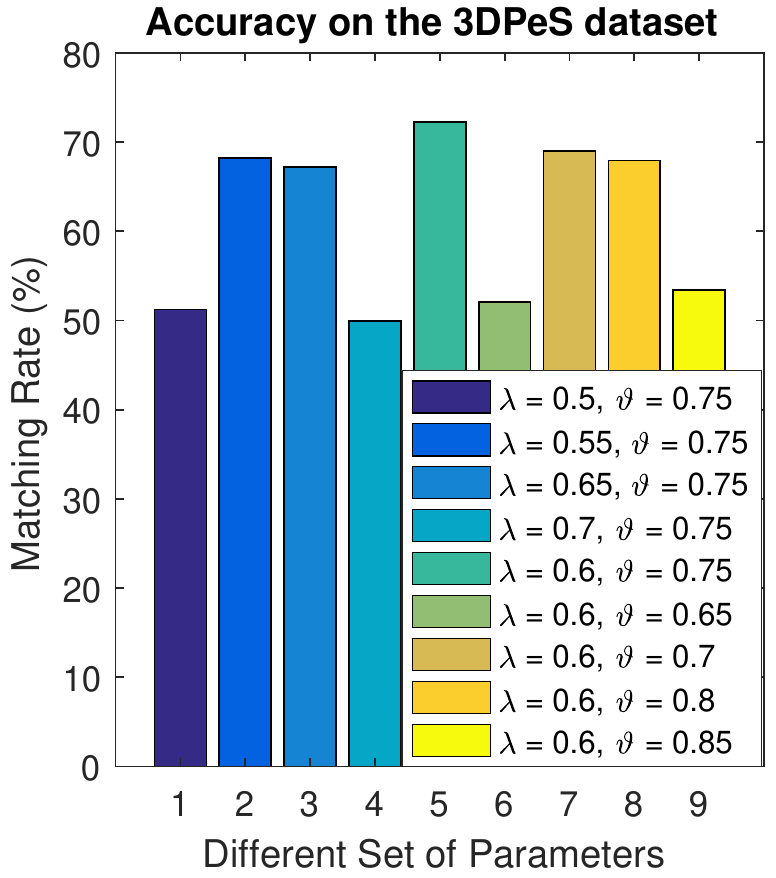} &
        \hspace{-0.5cm}
        \includegraphics[height = 3.9cm, width = 3.6cm]{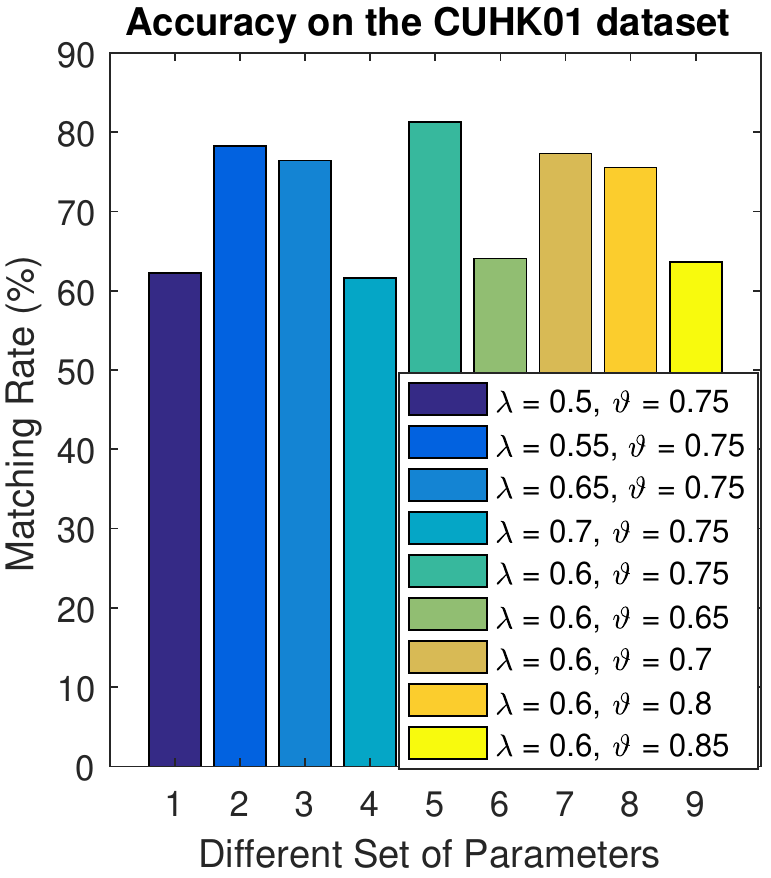}&
        \hspace{-0.5cm}
        \includegraphics[height = 3.9cm, width = 3.6cm]{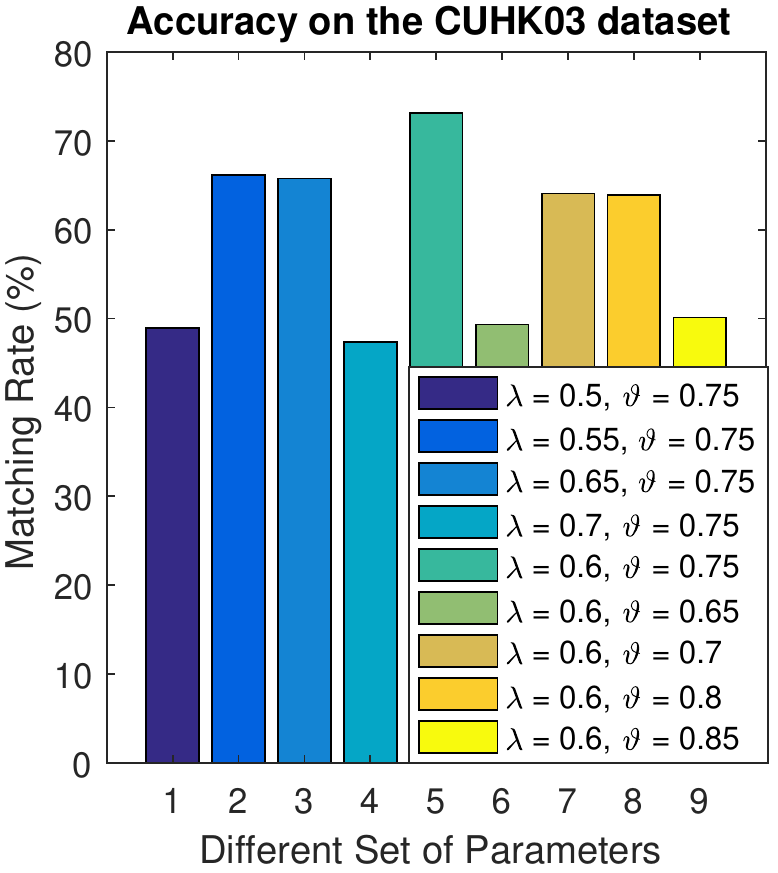}&
        \hspace{-0.5cm}
        \includegraphics[height = 3.9cm, width = 3.6cm]{SPL_analysis_Market1501.pdf}\\
    \end{tabular}
    \vspace{-0.3cm}
    \caption{Influences of different parameter settings to the final performance. Specially, the first column shows how the symmetric regularizer term effects the performance, in which our DSPL method get its best performance by setting $\mathcal{M} = 1.1$, $\zeta = 0.1$. The second column shows how the SPL strategy effects the performance, in which our DSPL method get the promising results by setting $\lambda = 0.6$ and $\vartheta = 0.75$ on the five benchmark datasets, respectively. }
    \label{fig_7}
\end{figure*}

\begin{table}[h]
\footnotesize
\caption{Matching rates of the SPL with different loss functions.}
\begin{center}
\label{tab_7}
\begin{tabular}{ c| c | c | c| c | c}
\hline
Metrics & VIPeR & 3DPeS & CUHK01 & CUHK03 & Market501\\
\hline
\hline
CL          & 42.35 & 54.81 & 67.61 & 51.43 & 57.61\\
CL+SPL      & \textcolor{red}{\bf{50.81}} & \textcolor{red}{\bf{62.49}} & \textcolor{red}{\bf{74.05}}
                         & \textcolor{red}{\bf{65.38}} & \textcolor{red}{\bf{66.28}}\\

\hline
TL              & 45.57 & 57.31 & 71.14 & 57.31 & 63.98\\
TL+SPL          & \textcolor{red}{\bf{55.22}} & \textcolor{red}{\bf{64.04}} & \textcolor{red}{\bf{77.38}}
                         & \textcolor{red}{\bf{69.43}} & \textcolor{red}{\bf{68.41}}\\
\hline
OurLoss     & 50.95 & 65.42 & 73.64 & 67.52 & 67.45\\
OurLoss+SPL & \textcolor{red}{\bf{56.32}} & \textcolor{red}{\bf{73.16}} & \textcolor{red}{\bf{81.33}}

                         & \textcolor{red}{\bf{73.16}} & \textcolor{red}{\bf{72.89}}\\
\hline
\end{tabular}
\end{center}
\end{table}

Secondly, we report two set of comparison results to evaluate the superiority of our part-based neural network and the effectiveness of the SPL strategy with different distance metrics. For fair comparison, we keep other parts the same when compare one specified part in all these experiments. In order to evaluate the superiority of our part-based network, we build another global network, in which we get rid of eight small convolutional layers in the local subnetwork and take two large convolutional layers to replace them. The comparison results are shown in Table~\ref{tab_6}, in which our part-based neural network outperforms the global neural network in Top 1 accuracy on all the five datasets. The reason may come from two aspect: 1) Different body parts have different importance in representing the person appearance~\cite{Ahmed_Jones_Marks:2015}, and the part-based neural network allows to learn different body parts discriminatively; 2) Dividing the feature maps into different parts is a kind of data augmentation, and the data augmentation is a common way to improve the network performance in deep learning community. Besides, we evaluate the effectiveness of the SPL strategy with three different distance metrics, namely the contrastive loss, the triplet loss and the proposed loss. The comparison results are shown in Table~\ref{tab_7}, in which the SPL strategy improves the baseline performance of different metrics on all the five datasets.  According to our experience, the reason of why the SPL works is due to the data distribution. As a fine-grained recognition task, person images usually gather together in feature space when the representation ability of the deep network is weak. At this time, easy samples are more beneficial to steadily enhance the representation ability of neural network. When the representation ability of deep model reaches a certain level, all the samples will be involved into the training process to boost the final performance.

{\bf Parameter influence} To the best of our knowledge, the margin parameter $\mathcal{M}$, the weight parameter $\zeta$ and the age parameters $\lambda, \vartheta$ have major effects to the final ranking performance in our DSPL method. The margin parameter $\mathcal{M}$ and weight parameter $\zeta$ jointly control the symmetric gradient back-propagation of the relative distance metric, and the age parameters $\lambda, \vartheta$ control the way of hard samples are involved into the training process. In the following, we give an empirical analysis of our method on the five datasets, respectively.

The results are shown in Fig.~\ref{fig_7}, in which our method achieves its best performance by setting $\mathcal{M} = 1.1, \zeta = 0.1$ and $\lambda = 0.6, \vartheta = 0.75$. We demonstrate the results from the following four points: 1) Small margin $\mathcal{M}$ will make the candidate positive and negative samples indistinguishable in the distance space; while a large margin will lead to the numerical instability problem. 2) Small weight $\zeta$ will weaken the symmetric constraint, which can lead to the asymmetric gradient back-propagation; while large weight will enhance the symmetric constraint, which is also harmful to computational stability. 3) The meaning of $\lambda$ is the current age of model, and small $\lambda$ will hinder the hard samples involved into the training process, while large $\lambda$ will lead the easy samples and hard samples indistinguishable in the training process. 4) The meaning of $\vartheta$ is the mature age of model, and small $\vartheta$ will lead the hard samples involve into the training process too early, while large $\vartheta$ will make only a small amount of hard samples are involved into the training process. Therefore, we choose a moderate $\mathcal{M}$, $\zeta$, $\lambda$ and $\vartheta$ for the high-quality performance of the symmetric gradient back-propagation and self-paced training process.

\begin{figure}[!htb]
\footnotesize
\centering
    \begin{tabular}{c}
        \hspace{-0.5cm}
        \includegraphics[height = 8cm, width = 9cm]{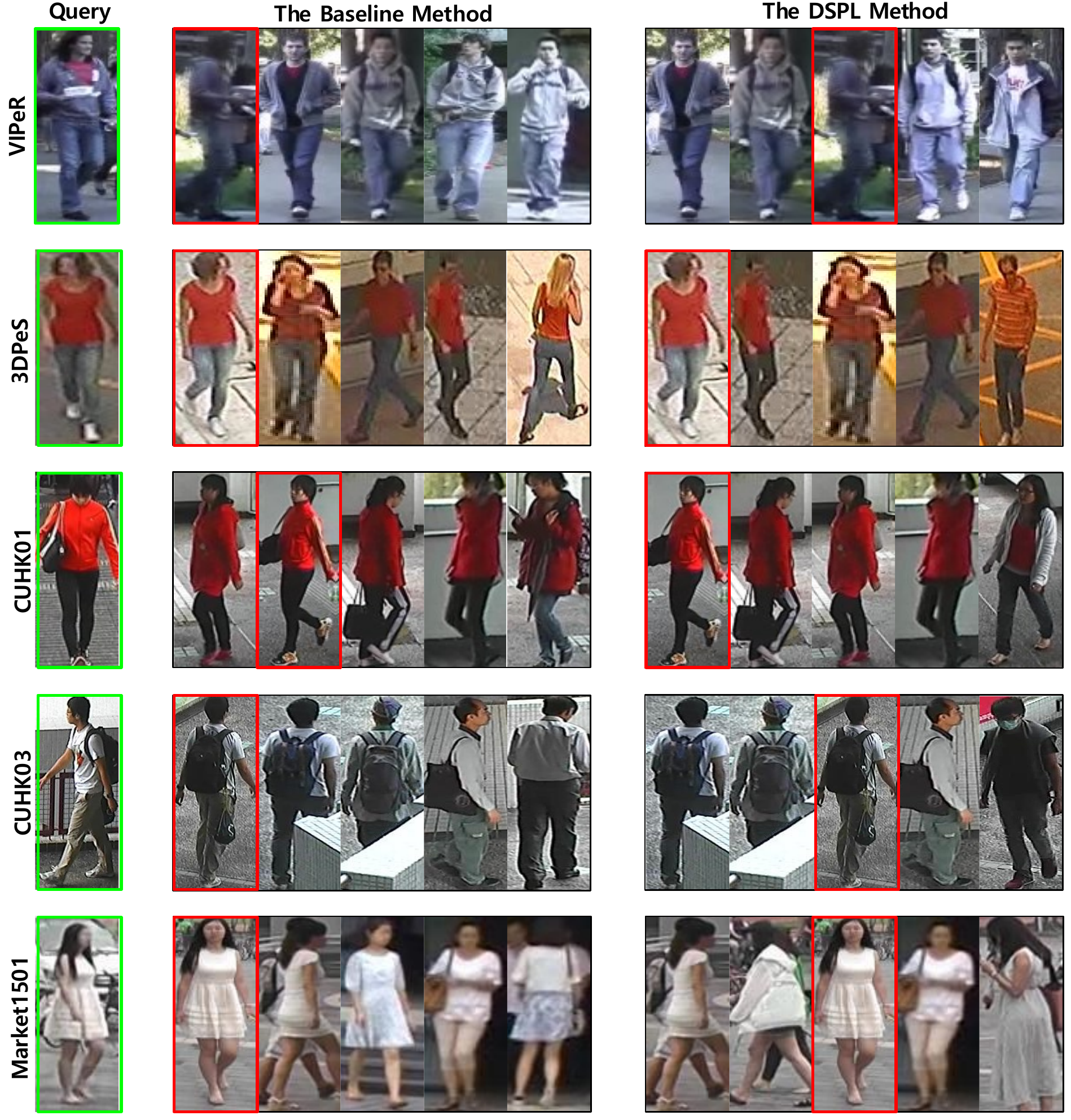}
    \end{tabular}
    \caption{The ranking results on the five benchmark datasets in single-shot evaluation, in which person in green rectangle denotes the query image and person in red rectangle represents the matched candidate. }
    \label{fig_8}
\end{figure}

{\bf Some ranking examples} To obtain more insight of the DSPL method, some ranking examples on the five benchmark datasets are shown in Fig.~\ref{fig_8}, in which person in green rectangle denotes the query image and person in red rectangle represents the matched candidate. For each dataset, we give the comparison ranking results of two methods, namely the baseline method and the DSPL
method, in the first and the second column respectively. From these examples, we can conclude our DSPL performs much better the baseline method. The main reason is that our DSPL method applies the SPL strategy and symmetric regularization in the training process, which have been effective to improve the person Re-ID performance. Compared with the ranking results on the CUHK01 and CUHK03 dataset, we find that our method can not always find the correct matches. In the future, we will strive to find an optimal saliency detection modular in our neural network, so as to further improve the ranking performance of cases as shown the CUHK01 and CUHK03 datasets.

\section{Conclusion}
\label{sec_conclusion}
In this paper, we propose a novel person re-identification method by incorporating the SPL strategy and symmetric regularizer to perform integrated feature learning and fusion in an end-to-end deep framework. In order to extract the stable and discriminative features, we build a part-based neural network, in which the features of different body parts are first discriminately learned in the lower convolutional layers and then fused in the higher fully connected layers. The output features are further fed into the relative similarity comparison metric to optimize the deep parameters in gradient back-propagation. By introducing the SPL strategy into the distance metric, the sides effects of noisy samples or outliers can be alleviated by using a soft polynomial regularizer to adaptively update the sample weights in each iteration. The asymmetric gradient back-propagation is revised by introducing the symmetric regularizer, therefore the intra-class distance is minimized and inter-class distance is maximized in each triplet unit. Extensive experimental results on the VIPeR, 3DPeS, CUHK01, CUHK03 and Market1501 datasets have shown that our method outperforms most of the state-of-the-art approaches in person re-identification.

{\small
\bibliographystyle{elsarticle-num}
\bibliography{SPL_journal}
}

\end{document}